\documentclass{article}

\usepackage{arxiv}

\usepackage[utf8]{inputenc}
\usepackage[T1]{fontenc}
\usepackage{hyperref}
\usepackage{url}
\usepackage{booktabs}
\usepackage{amsfonts}
\usepackage{amsmath}
\usepackage{nicefrac}
\usepackage{microtype}
\usepackage{graphicx}
\usepackage{natbib}
\usepackage{multirow}
\usepackage{algorithm}
\usepackage{algpseudocode}

\graphicspath{{figures/}}

\title{When Is a Learned Command Adapter Worth It?\\
Closed-Loop Identification and Counterfactual Auditing of Frozen Locomotion Policies}
\renewcommand{\shorttitle}{Auditing Learned Adapters for Frozen Policies}

\author{
  Zongtan Li \\
  \texttt{zongtanli2023@gmail.com} \\
}

\date{July 2026}

\begin{document}
\maketitle

\begin{abstract}
When a frozen, command-conditioned locomotion policy is deployed as a building
block, a common extension is a learned command adapter. We argue that this step
should be preceded by a diagnostic question: does the interface expose an
improvement that is both real at the same state and recoverable from
deployment-time features beyond a non-adaptive mixture of the same actions?
We formalize an \emph{adapter necessity audit} that separates three often
conflated quantities---global operating-point gain, same-state counterfactual
headroom, and state-allocation gain over a frequency-matched randomized
mixture---and requires both deployment gain over a cross-fitted fixed reference
and allocation gain before issuing a
\textsc{Go}/\textsc{No-Go}/\textsc{Abstain} decision under user-specified
practical-value and violation thresholds. Source-cluster resampling refits the
learner and evaluates it on out-of-bootstrap clusters. Closed-loop command-response
identification supplies optional decision features: from excitation rollouts, a
supervised model predicts situational motion, mechanical cost, and
terrain-relative posture (displacement $R^2\ge0.92$, posture $R^2\approx0.9$,
cost $R^2\approx0.75$--$0.8$). An archived Go2 scale-prefix diagnostic finds
$5.2\%$ same-state oracle headroom but only $0.55\%$ recovered allocation gain.
We then execute an action-aligned audit with direct, scale, heading, and yaw
interventions on twenty independent clusters for each of three Go2 query
distributions induced by direct control, VGCC, and MPC; uncertainty uses 200
full learner refits. With
$\delta_{\rm dep}=\delta_{\rm alloc}=1\%$ and $\kappa=5\%$, direct queries
return \textsc{No-Go}, whereas VGCC and MPC queries \textsc{Abstain}. VGCC has
the largest mean deployment gain ($1.34\%$), but its allocation lower bound is
only $0.09\%$ and its violation upper bound is $6.25\%$. A separate
twenty-cluster H1 audit using deployment-representative direct queries returns
\textsc{No-Go} (deployment-gain CI $[-0.81,0.71]\%$). A learner-level synthetic
control returns \textsc{Go} under the same rule. The primary contribution is a
counterfactual assessment of whether observable signal supports state-dependent
command adaptation.
\end{abstract}


\section{Introduction}
\label{sec:intro}

Strong learned locomotion policies are increasingly deployed as frozen,
command-conditioned building blocks: a task layer drives a pretrained
velocity-tracking policy toward a goal through its command interface, typically
with a hand-designed proportional command. In this regime the policy's weights,
reward, and training procedure are unavailable or impractical to modify; one can
only observe the policy acting and choose the next command. A common response is
to build a learned adapter. We consider a logically prior question:
does the interface expose an improvement that is real at the same state, and
recoverable from deployment-time features beyond merely changing the empirical
composition of commands?

Controller rankings alone cannot answer that question. Beating a proportional
baseline may reflect a better global operating point rather than useful
state-conditioned allocation; an inability to outperform a tuned fixed scale may mean either
that no local headroom exists or that the available representation cannot recover
it. We therefore introduce an \emph{adapter necessity audit} that separates three
quantities: (i)~global operating-point gain from choosing one fixed intervention;
(ii)~same-state counterfactual headroom of an oracle that observes realized
outcomes; and (iii)~state-allocation gain of a cross-fitted selector over a
randomized mixture with identical action frequencies. A deployment decision
additionally requires improvement over a cross-fitted fixed reference. Whole
source clusters are resampled and the learner is refitted; user-specified
deployment-value, allocation-value, and violation thresholds then determine
\textsc{Go}, \textsc{No-Go}, or \textsc{Abstain}. Analyses without the declared
cluster count and refitting are descriptive. Abstention is represented
explicitly: an underpowered audit is not evidence that an adapter is unnecessary.

To supply optional decision features---and as a reusable measurement of the
closed loop---we identify the frozen policy--robot response from excitation
rollouts. A supervised model predicts, for any candidate command in the current
situation, short-horizon motion, mechanical cost, and terrain-relative posture.
Modeling the closed loop at the three-dimensional command interface, rather than
action-level dynamics, makes identification a tractable supervised problem on the
policy's own rollouts. Average prediction accuracy is high (displacement
$R^2\ge0.92$; posture $R^2\approx0.9$; cost $R^2\approx0.75$--$0.8$), and
terrain-relative posture labels are necessary on rough ground. Accuracy alone,
however, does not certify ranking value: the audit tests whether these
predictions improve counterfactual selection beyond the observation itself.

As a concrete adapter under that audit we study
\textbf{Viability-Gated Command Compensation} (VGCC): a training-free controller
that scores a structured candidate set around the proportional command and
applies a bounded correction toward the lowest predicted-cost eligible candidate,
otherwise reverting to direct control. Across three embodiments it lowers running cost
and mechanical work relative to the proportional baseline, yet neither VGCC nor a
heavier sampling MPC outperforms a tuned fixed-scaling frontier. An archived
scale-prefix analysis first diagnoses the scalar family shared with MPC. An episode-level oracle over
three scales improves on the best global scale
by only $1.0\%$ on Go2 (the $1.9\%$ H1 Stage~1 screen is descriptive). Exact
same-state replay on ten source clusters finds
$5.2\%$ local work headroom under success and a $10\%$ time budget, so local
opportunity exists. Recoverability is limited: a treatment-effect ensemble realizes
only $0.55\%$ beyond a frequency-matched mixture, while its candidate
treatment-effect error ($11.6\%$) exceeds the signal it must rank. The
confirmatory audit then adds heading and yaw interventions and evaluates twenty
source clusters per Go2 query distribution. At the prespecified $1\%$ deployment
and allocation thresholds, direct queries yield \textsc{No-Go}, while VGCC and
MPC queries yield \textsc{Abstain}; a deployment-representative twenty-cluster
H1 audit yields \textsc{No-Go}. Comparator disagreement is itself diagnostic:
on H1, the selector appears to improve slightly over direct control but loses to
the cross-fitted scale-$0.90$ reference. Thus the richer audit does not certify
a deployment-ready adapter, while distinguishing inadequate state--action
matching, an inadequate fixed reference, and unresolved uncertainty.

\textbf{Contributions.}
(1)~\emph{Robotics instantiation of personalization-value auditing.} Building on
tests of whether individualized rules beat a constant intervention, a protocol
for closed-loop adapters that separates global operating-point gain, same-state
counterfactual headroom, and recoverable
state-allocation gain over a matched randomized mixture, and deployment gain
over a cross-fitted fixed reference, with an explicit three-way decision under
separate practical-value and violation thresholds.
(2)~\emph{Closed-loop command-response identification.} A data-efficient model
of situational motion, mechanical cost, and terrain-relative posture for a
frozen command-conditioned policy, with terrain-relative posture labels shown
to be necessary and with identification treated as an optional feature source
rather than an assumed prerequisite for the audit.
(3)~\emph{Action-aligned empirical diagnosis.} On standard rough-terrain
reaching, the archived scale diagnostic exposes $5.2\%$ local headroom but only
$0.55\%$ recovered allocation gain. The confirmatory analysis adds heading and
yaw interventions, twenty-cluster direct/VGCC/MPC query distributions, a
cross-fitted fixed comparator, and 200 learner refits. It produces no real-domain
\textsc{Go}: direct Go2 and deployment-representative H1 queries yield
\textsc{No-Go} at the $1\%$ thresholds, while VGCC and MPC queries
\textsc{Abstain}. The two comparators prevent gain from action-frequency changes
or a weak direct baseline from being misattributed to personalization.
Observable, hidden, and learner-level positive controls calibrate the decision
channels.

\section{Related Work}
\label{sec:related}

\textbf{Improving pretrained controllers without retraining.}
Residual policy learning adds action-space corrections but requires reward access
and further interaction~\citep{silver2018residual,johannink2019residual}.
RMA~\citep{kumar2021rma} and multiplicity-of-behavior
training~\citep{margolis2022walk} expose adaptation at training time.
Massively parallel training and terrain curricula improve the base policy during
training~\citep{rudin2022learning,makoviychuk2021isaac,mittal2023orbit,lee2020learning,miki2022learning}.
We study the complementary setting in which the policy is frozen and adaptation
is restricted to its command interface. The central question is whether the
observable evidence justifies a learned adapter.
Interface-level optimization over a fixed low-level controller is successful in
eco-driving~\citep{sciarretta2015ecodriving} and reference
governors~\citep{garone2017reference}; our audit asks when the analogous claim
holds for efficiency over a velocity-commanded locomotion policy.

\textbf{Closed-loop models for command-level control.}
Model-based RL typically learns action-level
dynamics~\citep{deisenroth2011pilco,chua2018deep,janner2019trust,hafner2020dream,hansen2022temporal,nagabandi2018neural}.
When a competent policy already exists, modeling the closed loop at the command
interface collapses input dimensionality and restricts decisions to commands the
policy understands. Related instantiations include reference governors and
predictive safety filters on prestabilized
loops~\citep{garone2017reference,wabersich2021predictive}, ABS searching twist
commands against a co-trained reach-avoid value~\citep{he2024agile}, and
Koopman identification of an RL-controlled biped for safe
navigation~\citep{kim2024koopman}. Feedforward/internal-model
control~\citep{astrom2008feedback,francis1976internal} and sampling
MPC~\citep{mayne2000constrained,williams2018information,kabzan2019learning,dicarlo2018dynamic}
supply the relevant control formulations. VGCC is an instance of this
family---gated one-step compensation from an identified response model---and
serves as the empirical case study. Unlike prior systems that treat identification as a means to control, we
treat it as a falsifiable feature source: we ask whether its predictions improve
counterfactual ranking beyond the observation and beyond a frequency-matched
non-adaptive mixture.

\textbf{Treatment choice and policy value.}
Learning a command adapter from same-state interventions is a treatment
assignment problem: context is observed before choosing an intervention, and the
goal is policy value rather than marginal prediction accuracy. Empirical welfare
maximization~\citep{kitagawa2018treatment}, efficient policy
learning~\citep{athey2021policy}, and counterfactual risk
minimization~\citep{swaminathan2015crm} formalize this distinction.
More directly, \citet{cai2020validation} test whether an individualized decision
rule improves on a naive fixed treatment, and the K-fold personalization test of
\citet{li2026personalization} asks whether a personalized policy surpasses the
best single intervention while controlling selection-induced bias. Accordingly,
our contribution is neither the general proposition that personalization should
be tested against a constant policy nor a new asymptotic test.
Our setting has deterministic exact-replay potential outcomes rather than
unknown propensities, but adds robotics-specific requirements: task success/time
eligibility, closed-loop query-state distributions, and comparison of a fitted
selector with a randomized policy having its \emph{own} intervention frequencies.
The best-constant comparator in personalization testing asks whether any
contextual rule can beat one global action. Our matched-frequency comparator asks
a different attribution question about a particular learned selector: holding its
action composition fixed, is its state--action matching valuable? Algebraically
it removes gains obtainable by changing marginal command frequencies and retains
the context--outcome covariance. It is a diagnostic estimand, not a replacement
for the best-constant hypothesis test.

\textbf{Options and behavior archives.}
Temporal abstraction builds reusable behavior
libraries~\citep{sutton1999options,bacon2017option,eysenbach2019diversity,sharma2020dynamics,peng2022ase,mouret2015illuminating}.
DADS plans over skill transition models co-trained with the
skills~\citep{sharma2020dynamics}; we instead identify an already-trained task
policy. A retrieval-based archive variant is characterized in
Appendix~\ref{app:retrieval} as coverage-limited relative to the parametric
response model. We intervene at the command interface because raw chunk replay
destabilizes contact-rich locomotion~\citep{zhao2023learning}.

\section{Method}
\label{sec:method}

The method has three parts. The primary object is an \emph{adapter necessity audit}
that separates available counterfactual headroom from the value a
state-conditioned selector recovers beyond a frequency-matched randomized policy
(Section~\ref{sec:audit}). Closed-loop command-response identification
(Section~\ref{sec:model}) supplies optional decision features for that audit and
for any compensator built on the same interface. VGCC
(Section~\ref{sec:vgcc}) is one concrete adapter evaluated by the audit: a bounded,
gate-certified correction toward the lowest-cost member of a structured candidate
set. Identification is not assumed to be necessary for a positive audit decision;
the audit measures its incremental contribution as a feature source.

\textbf{Design requirements.}
\emph{R1, observation only}: weights, reward, and training procedure are
unavailable; everything must come from observing the policy act.
\emph{R2, non-degradation}: any deployed adapter must revert to the pretrained
policy exactly when its intervention is uncertain.
\emph{R3, embodiment-light}: the method does not require robot-specific knowledge
beyond quantities identified from rollouts; posture thresholds may depend on
morphology, while objectives and model architecture remain shared.
These constraints shape the interface (Section~\ref{sec:setting}), the response
model, the audit protocol, and the VGCC case study.

\subsection{Compensation Interface}
\label{sec:setting}

Let $\pi_\theta$ be a frozen, pretrained continuous-control policy mapping
observations $o_t$ and a command $c$ to low-level actions.
In our locomotion instantiation $c = (v_x, v_y, \omega_z)$ is the body-frame
velocity command, but the framework requires only a command interface
through which behavior can be steered, a property shared by learned controllers
deployed as task-level building blocks~\citep{margolis2022walk,kumar2021rma}.

\textbf{Scope.}
Three properties delimit where the framework applies.
\emph{(i)~Command interface}: an unconditioned single-task policy exposes nothing
to invert over.
\emph{(ii)~Command redundancy}: several commands must make comparable progress at
different cost; a fully determined command admits no efficiency gain.
\emph{(iii)~Fast closed-loop response}: command consequences must be visible within
a short prediction window (locomotion, end-effector servoing, vehicle tracking),
not sparse long-horizon credit assignment.
Section~\ref{sec:limitations} returns to what lies outside these bounds.

We intervene at the command rather than the action for three reasons aligned with
R1--R3. Action residuals require reward and further interaction; a command
correction can be computed from an identified model. Command corrections leave
$\pi_\theta$ in control of contacts, but still need explicit fallback and posture
gating (R2). The command interface is shared across embodiments (R3).

A task supplies a target; throughout the method, $g$ denotes that target expressed
in the robot's body frame at the current step, and the \emph{direct} controller
computes a proportional command
\begin{equation}
c_{\text{goal}} \;=\; \bigl(\operatorname{clip}_{[-v_x^{\max},v_x^{\max}]}(k_v x_{\text{local}}),\;
\operatorname{clip}_{[-v_y^{\max},v_y^{\max}]}(k_v y_{\text{local}}),\;
\operatorname{clip}_{[-\omega^{\max},\omega^{\max}]}(k_\psi \operatorname{atan2}(y_{\text{local}},x_{\text{local}}))\bigr),
\label{eq:cgoal}
\end{equation}
which defines the uncompensated baseline: a strong pretrained policy commanded
toward the goal. Such proportional commands are a standard task layer for
velocity-conditioned policies; they saturate far from the goal and induce
energy-intensive in-place stepping near it. Compensation replaces
$c_{\text{goal}}$ when---and only when---a corrected command is trusted.

\subsection{Command-Response Model}
\label{sec:model}

Adapters and audits both need predictions of what a candidate command would do.
We model the \emph{closed loop} of policy and robot at the command interface,
not action-level dynamics~\citep{nagabandi2018neural}: the policy already handles
contacts, and re-modeling them would duplicate that work at far higher
dimensionality. The closed-loop view keeps the input three-dimensional and makes
identification a supervised problem on the policy's own rollouts (R1, R3).

\textbf{Excitation data.}
We override the converged policy with commands sampled throughout its admissible
command space, including low-magnitude commands near the goal. This is persistent
excitation applied to the policy--robot closed loop.
No quality filtering is applied: stumbles, slips, and low-posture episodes are
retained so that the response model covers low-performance regions. Restricting
training to high-quality segments yields inadequate predictions in precisely the
regions for which the viability gate is intended (Section~\ref{sec:results-model}).

\textbf{Response targets.}
The outputs instantiate a general three-channel schema that any instantiation of
the framework must fill: a \emph{progress} statistic sufficient to score task
advancement, the \emph{cost} being minimized, and the \emph{viability} quantity
the gate guards.
For locomotion, constant-command windows provide body-frame displacement
$(\Delta x, \Delta y)$ and yaw change $\Delta\psi$ for progress; a mechanical cost
for the objective; and minimum and mean terrain-relative posture height for
viability.
We use two cost channels of increasing physical fidelity. The first is a
dimensionless \emph{actuation-effort proxy}
$\bar{a}=\operatorname{mean}_{t,j}(|a_{t,j}\dot q_{t,j}|)$, formed from the public
policy action $a$; because $a$ is not a measured motor torque, this proxy has no
physical energy unit and is treated as an optimization objective, not an energy
measurement. The second is a torque-derived mechanical power
$\bar P=\operatorname{mean}_{t,j}(|\tau_{t,j}\dot q_{t,j}|)$, where $\tau_{t,j}$ is
the simulator-applied joint torque; this is a simulated mechanical quantity, still
not battery energy, but the physically meaningful channel used to audit whether
reducing the proxy reduces mechanical work.
A manipulation instantiation would substitute end-effector displacement, an
appropriate actuation cost, and a grasp-stability or joint-limit margin.
Posture height is recovered from the policy's own height-scan observation (mean
scan return plus sensor offset), not from world coordinates. This prevents ground
elevation from being confounded with posture; its empirical value is reported with
the response-model results.

\textbf{Model.}
The response model is a mapping
\begin{equation}
f_\phi : (o_t, c) \;\longmapsto\; \bigl(\widehat{\Delta x}, \widehat{\Delta y}, \widehat{\Delta\psi}, \hat C, \hat{h}_{\min}, \hat{h}_{\text{mean}}\bigr),
\label{eq:responsemodel}
\end{equation}
where hats mark quantities \emph{predicted} by the model, plain symbols are
measured, and $\hat C$ is the chosen cost channel. The model receives the full
policy observation with its command component replaced by the candidate
command; architecture, training, and split details are given in
Appendix~\ref{app:impl}.
Conditioning on the full observation makes the predictions context-dependent:
the same candidate command may have different predicted costs on
different terrain and from different gait states, which is precisely the
information a fixed command-shaping heuristic cannot express.

\subsection{Adapter Necessity Audit}
\label{sec:audit}

\begin{figure}[t]
\centering
\includegraphics[width=\linewidth]{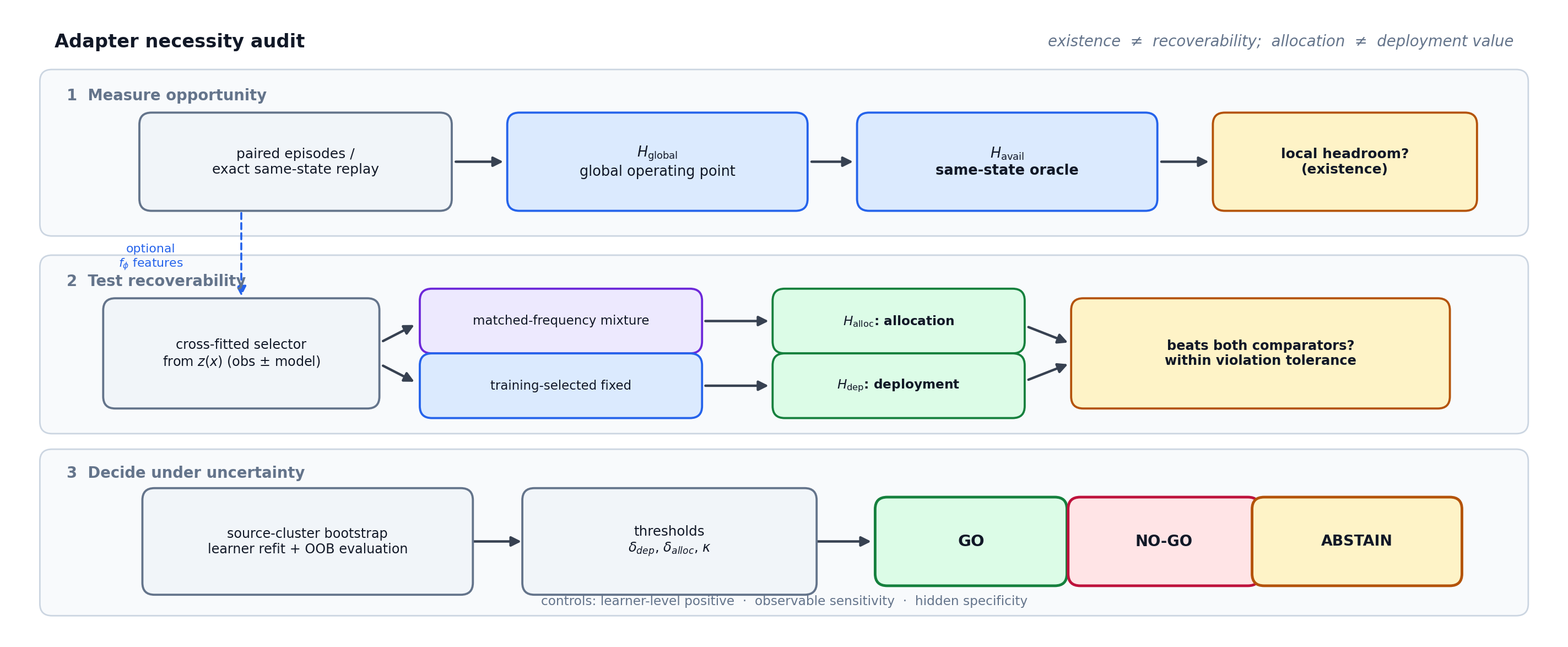}
\caption{Adapter necessity audit. Stage~1 measures opportunity
($H_{\rm global}$, $H_{\rm avail}$). Stage~2 isolates state-allocation gain
$H_{\rm alloc}$ over a frequency-matched mixture and deployment gain
$H_{\rm dep}$ over a cross-fitted fixed reference. Stage~3 maps source-cluster
uncertainty into \textsc{Go}/\textsc{No-Go}/\textsc{Abstain} under thresholds
$(\delta_{\rm dep},\delta_{\rm alloc},\kappa)$, with a learner-level positive
control and observable/hidden controls for sensitivity and specificity.
Identified-model features $f_\phi$ are optional inputs to Stage~2.}
\label{fig:audit}
\end{figure}

Average response accuracy does not determine whether a learned adapter is worth
deploying (Figure~\ref{fig:audit}). The decision depends on three distinct
quantities: whether alternative commands can help at the same state, whether
their treatment effects are predictable from deployment-time information, and
whether state-conditioned allocation beats issuing the same commands at the same
frequencies without conditioning. We formalize these quantities before introducing
any particular adapter.

Let $x$ denote the observable query context (observation and task target),
$a_0$ the direct command, and $\mathcal A$ a finite intervention set. Let
$Y_x(a)=(W_x(a),T_x(a),S_x(a))$ be the potential mechanical work, completion
time, and success after taking intervention $a$ and following a common
continuation. For a time budget $\rho$, candidate $a$ is eligible when
\begin{equation}
E_x(a)=\mathbf{1}\!\left[S_x(a)\ge S_x(a_0),\;
T_x(a)\le\rho T_x(a_0)\right]=1 .
\label{eq:audit-eligible}
\end{equation}
In simulation we observe these potential outcomes by exact start-state replay;
without reset access they must instead be estimated from randomized or logged
interventions, a limitation we return to in Section~\ref{sec:limitations}.

\textbf{Available headroom.}
The best fixed action
$a^{\rm g}=\arg\min_{a\in\mathcal A}\mathbb E_x[W_x(a)]$ (subject to the
declared aggregate constraints) measures global operating-point gain
$H_{\rm global}=1-\mathbb E W_x(a^{\rm g})/\mathbb E W_x(a_0)$.
The same-state oracle $\pi^\star(x)=\arg\min_{a:E_x(a)=1}W_x(a)$ instead measures
the full local opportunity present in $\mathcal A$:
\begin{equation}
H_{\rm avail}=1-\frac{\mathbb E_x[W_x(\pi^\star(x))]}
{\mathbb E_x[W_x(a_0)]}.
\label{eq:havailable}
\end{equation}
This is an oracle diagnostic, not a deployable controller.

\textbf{Recovered and allocation value.}
A selector $\hat\pi(z(x))$ is cross-fitted across independent source seeds from
deployment features $z$; these may contain only observations or append identified
response predictions. Its total recovered gain is
$H_{\rm total}=1-\mathbb E W_x(\hat\pi(x))/\mathbb E W_x(a_0)$.
Total gain alone is insufficient because any selector can improve simply by
issuing a globally lower-cost scale more often. We therefore construct
$\mu_{\hat\pi}$, which draws actions independently of $x$ with exactly the
marginal frequencies
$p_a=\Pr_x[\hat\pi(x)=a]$ of the learned selector. The state-allocation value is
\begin{equation}
H_{\rm alloc}=1-\frac{\mathbb E_x[W_x(\hat\pi(x))]}
{\mathbb E_{x,a\sim p}[W_x(a)]}.
\label{eq:halloc}
\end{equation}
Only $H_{\rm alloc}>0$ is evidence that conditioning commands on context adds
value beyond a matched non-adaptive mixture. We separately report the realized
constraint-violation rate
$q=\Pr_x[E_x(\hat\pi(x))=0]$.
The numerator of this attribution can also be written
\begin{equation}
\mathbb E_{x,a\sim p}W_x(a)-\mathbb E_xW_x(\hat\pi(x))
=-\sum_{a\in\mathcal A}\operatorname{Cov}_x
\!\left(\mathbf 1\{\hat\pi(x)=a\},W_x(a)\right).
\label{eq:matched-covariance}
\end{equation}
Thus the comparator removes marginal action-composition effects: it is
positive only when the selector assigns an action disproportionately to contexts
where that action's outcome is better. Unlike comparison with the best constant
action, it distinguishes context-dependent allocation from total gain attributable
to more frequent use of a globally low-cost command.

Allocation value is an attribution quantity, not by itself a deployment
criterion. In held-out cluster $k$, let $a^{\rm g,-k}$ be the lowest-work fixed
action satisfying the declared violation tolerance on the training clusters
only. Its held-out deployment comparison is
\begin{equation}
H_{\rm dep}=1-\frac{\mathbb E_x[W_x(\hat\pi^{-k}(x))]}
{\mathbb E_x[W_x(a^{\rm g,-k})]}.
\label{eq:hdeployment}
\end{equation}
Cross-fitting prevents held-out outcomes from selecting the fixed reference.
Positive $H_{\rm alloc}$ therefore establishes state-dependent attribution,
whereas positive $H_{\rm dep}$ establishes improvement over a deployable fixed
policy. Both are required for a deployment recommendation.

\textbf{Three-way deployment decision.}
Because a statistically positive but operationally negligible gain does not
justify an adapter, the deployer specifies minimum practical values
$\delta_{\rm dep}$ and $\delta_{\rm alloc}$ and maximum violation rate $\kappa$.
Uncertainty is estimated by resampling whole source clusters, refitting the
learner, and evaluating on out-of-bootstrap clusters. Let $[L_D,U_D]$,
$[L_A,U_A]$, and $[L_q,U_q]$ denote the resulting intervals for $H_{\rm dep}$,
$H_{\rm alloc}$, and $q$. The audit returns
\begin{equation}
\begin{aligned}
\textsc{Go}&:\ L_D>\delta_{\rm dep}\ \land\
L_A>\delta_{\rm alloc}\ \land\ U_q\le\kappa,\\
\textsc{No-Go}&:\ U_D<\delta_{\rm dep}\ \lor\
U_A<\delta_{\rm alloc}\ \lor\ L_q>\kappa,\\
\textsc{Abstain}&:\ \text{otherwise}.
\end{aligned}
\label{eq:audit-decision}
\end{equation}
The abstention region is essential: an underpowered audit is not evidence that
an adapter is unnecessary. Confirmatory reporting requires at least twenty
independent deployment clusters and learner-refit uncertainty; otherwise the
quantities are descriptive and no formal decision is assigned. We report
threshold sensitivity rather than selecting practical-value thresholds after
observing the result.

\textbf{Reusable protocol.}
Before data collection, an audit instantiation declares
$(\mathcal A,W,E,\delta_{\rm dep},\delta_{\rm alloc},\kappa)$, the independent deployment unit to be
resampled, and a minimum confirmatory cluster count. The intervention set must be
aligned with the action families addressed by the resulting inference; a
scale-only audit supports conclusions about scale selection but not heading or
yaw selection. It then (i) measures the fixed operating-point frontier,
(ii) obtains paired potential outcomes on representative query contexts,
(iii) cross-fits candidate selectors from deployment-available representations
and evaluates each against its cross-fitted fixed reference and
frequency-matched mixture, and (iv) applies
Eq.~\ref{eq:audit-decision} together with observable- and hidden-signal controls.
The protocol applies beyond locomotion when the intervention, outcome, and
independent deployment unit can be specified; exact replay, mechanical work, and
command interventions are the instantiation used here.

\subsection{Viability-Gated Compensation}
\label{sec:vgcc}

\begin{figure}[t]
\centering
\includegraphics[width=\linewidth]{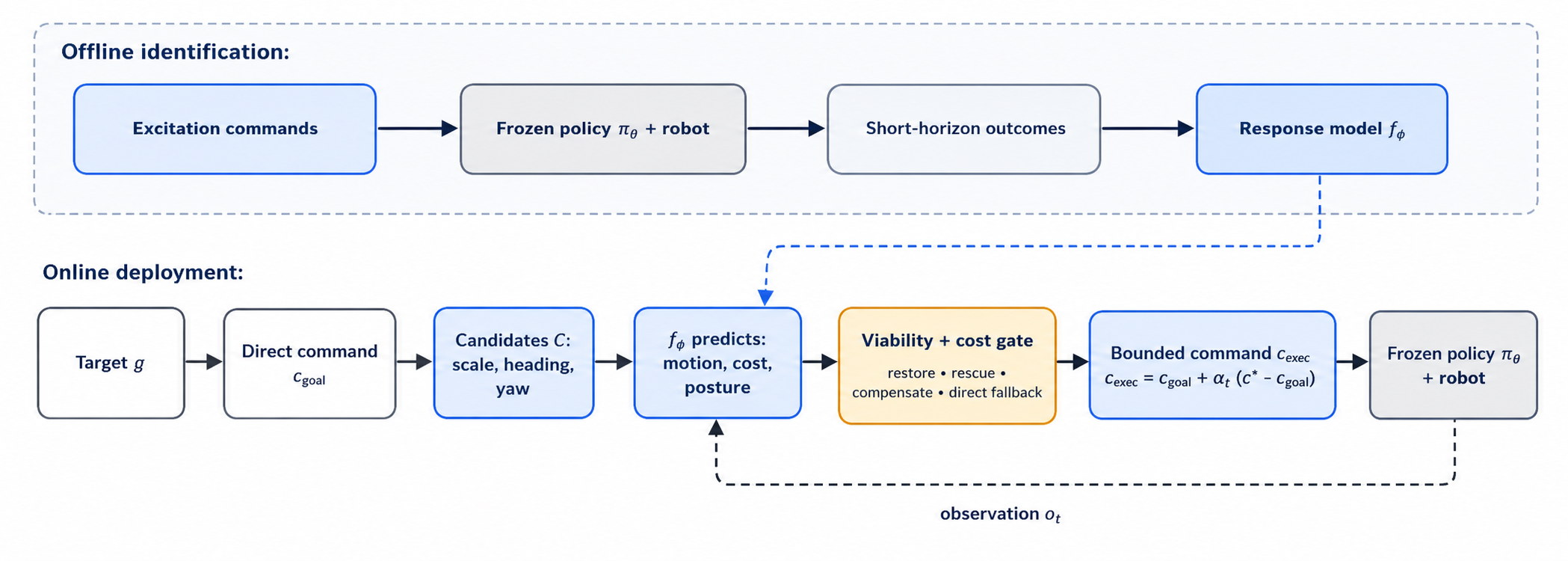}
\caption{VGCC overview (case study under the audit). Offline excitation rollouts
train an observation-conditioned response model. Online, the model scores a
structured candidate set around the direct command; a priority-ordered gate
applies a bounded correction only when progress, posture, and cost-margin checks
pass, and otherwise uses exact direct fallback.}
\label{fig:method}
\end{figure}

VGCC scores a structured candidate set built around the direct command:
uniform scalings of $c_{\text{goal}}$ (factors $0.4$ to $1.2$), yaw-rate offsets,
and a structured polar grid of absolute speeds at heading offsets around the target
direction, about $70$ candidates in total, all clipped to the admissible command
box (exact values in Appendix~\ref{app:impl}). One batched forward pass of
$f_\phi$ predicts every candidate's response. The set is wide enough to contain
slower, redirected, and re-timed variants of the direct command, but each
candidate is still a single command held for one decision interval, not a plan.

\textbf{Scoring and execution.}
The gate scores raw candidates; what it executes is a bounded correction toward
the selected candidate $c^\ast$,
$c_{\text{exec}}=c_{\text{goal}}+\alpha_t\,(c^\ast-c_{\text{goal}})$, with gain
$\alpha_t=\alpha\operatorname{clip}(d/d_{\text{anneal}},0,1)$ that anneals to
zero as the remaining distance $d$ shrinks, so corrections fade out on final
approach. The executed command therefore always lies between the strong direct
command and a model-certified candidate. Scoring the raw candidate (not the
blend) is a simplification whose error is limited by the bounded gain.

\textbf{Progress floor as a slowdown bound.}
Predicted target progress is
\begin{equation}
\hat{G}(c) \;=\; \lVert g \rVert \;-\; \bigl\lVert g - (\widehat{\Delta x}(c), \widehat{\Delta y}(c)) \bigr\rVert .
\end{equation}
A candidate is eligible when it preserves most of direct progress and posture:
\begin{equation}
\hat G(c)\ge\beta\,\hat G(c_d),\qquad
\hat h_{\min}(c)\ge h_{\rm floor},\qquad
\hat h_{\rm mean}(c)\ge h_{\rm now}-\delta,
\label{eq:vgc-eligible}
\end{equation}
with $c_d=c_{\text{goal}}$. Because the horizon has fixed duration, $\beta=0.9$
caps instantaneous slowdown at approximately $10\%$ and excludes unconstrained
reductions in speed; the corresponding ablation quantifies this effect
(Section~\ref{sec:results-ablation}).

\textbf{Cost objective and margin.}
Among eligible candidates, VGCC selects
$c^\ast=\arg\min_{c\,\text{eligible}}\hat C(c)$ and intervenes only if
\begin{equation}
\hat C(c^\ast) < (1-\epsilon)\,\hat C(c_d).
\label{eq:vgc-cost}
\end{equation}
The progress floor bounds slowdown; $\epsilon$ rejects model noise. Normalizing
cost by predicted progress performs worse in a paired ablation
(Appendix~\ref{app:ablation}): with the floor already present, the extra
normalization over-suppresses compensation.

\textbf{Posture rescue and restore.}
Two reflexes complete the controller. If the model predicts that even the direct
command will breach the posture floor within the horizon, VGCC instead selects,
among candidates predicted to make non-negative progress, the one with the
highest predicted minimum height (\emph{rescue}); the bounded correction then
pulls the executed command toward it. If measured posture has already fallen
below the operating band, VGCC suspends compensation and issues the nominal
direct command until height recovers (\emph{restore}). Both reflexes default
toward the direct controller, preserving R2. The response model therefore
restricts candidate selection rather than defining an unconstrained optimizer.
All scalar parameters are selected on development seeds and fixed before confirmation;
Appendix~\ref{app:impl} lists their values.

Algorithm~\ref{alg:vgcc} in Appendix~\ref{app:impl} summarizes the procedure. The
one-step comparison reflects the local prediction horizon of $f_\phi$
(Section~\ref{sec:results-model}); the margin $\epsilon$ suppresses interventions
when predicted rankings are insufficiently separated. Section~\ref{sec:results-headroom}
evaluates whether average predictive accuracy translates into counterfactual
ranking performance.

\section{Experiments}
\label{sec:results}

Experiments answer three questions in the order needed by the audit.
\textbf{Q1}: how accurately can the closed-loop command response be identified
(Section~\ref{sec:results-model})?
\textbf{Q2}: as a case-study adapter, how does VGCC compare with proportional
control, tuned fixed scaling, a reactive governor, and sampling MPC
(Sections~\ref{sec:results-main}--\ref{sec:results-baselines}, with coverage in
Section~\ref{sec:results-coverage})?
\textbf{Q3 (central)}: how much same-state headroom exists, how much is
recoverable beyond a frequency-matched mixture, and what deployment decision does
the audit return (Section~\ref{sec:results-headroom})?
Q1 establishes the optional feature source; Q2 places concrete adapters on the
efficiency frontier; Q3 is the paper's main claim---existence versus
recoverability versus allocation.

\subsection{Setup}
\label{sec:setup}

\begin{figure}[t]
\centering
\includegraphics[width=\linewidth]{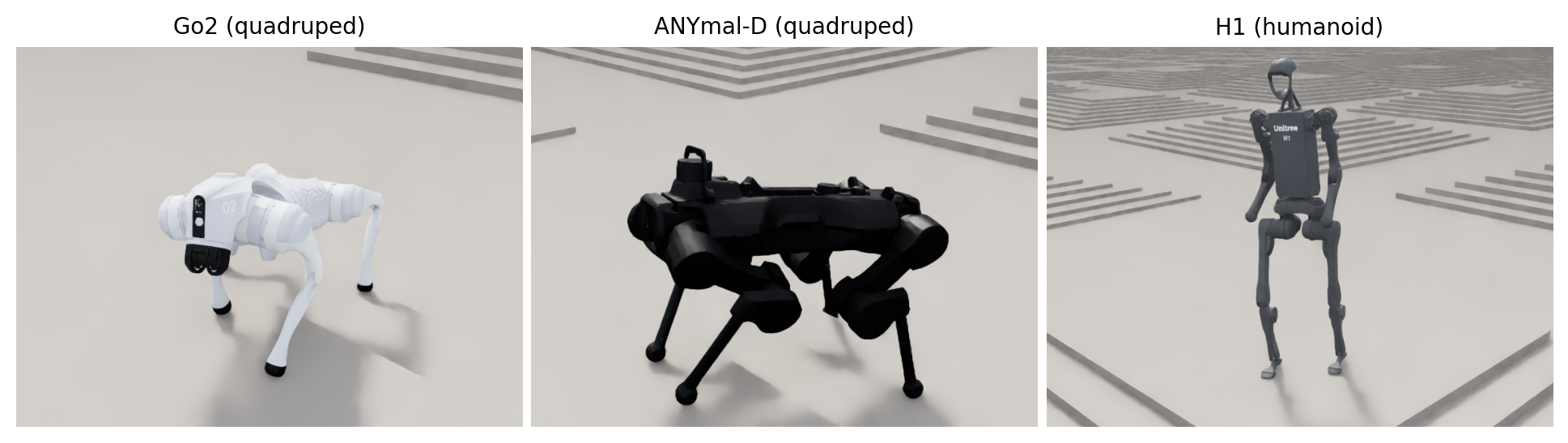}
\caption{The three embodiments in their Isaac Lab rough-terrain velocity
environments: the Go2 and ANYmal-D quadrupeds and the H1 humanoid. Each is driven
by its publicly distributed pretrained RSL-RL policy, which is the policy VGCC
compensates; VGCC changes only the command that policy receives, never its weights.
The two quadrupeds share every gate parameter unchanged; the humanoid differs only
in its posture thresholds.}
\label{fig:setup}
\end{figure}

\textbf{Environments and robots.}
All experiments use Isaac Lab~\citep{mittal2023orbit} rough-terrain velocity
environments with three embodiments (Figure~\ref{fig:setup}):
\texttt{Isaac-Velocity-Rough-Unitree-Go2-v0}
and \texttt{Isaac-Velocity-Rough-Anymal-D-v0} (quadrupeds) and
\texttt{Isaac-Velocity-Rough-H1-v0} (humanoid), with the publicly distributed
pretrained RSL-RL checkpoints~\citep{rudin2022learning,schwarke2025rslrl} as the
compensated policies.
The task is target reaching: move the base to a target $g$ specified relative to
the start pose.
Success requires no early termination, final XY distance below $0.3$\,m, and
\emph{final height} above $70\%$ of spawn height. Final height is a terminal
posture check, so that an episode that reaches the target crouched, collapsed, or
mid-fall does not count as a success. Because the terrain is procedurally rough,
final height reflects both posture and the ground elevation where the robot stops,
which is why we report it alongside, rather than inside, the cost and distance
metrics.

\textbf{Metrics.}
We report success, final distance, final height, and how often compensation was
active. The optimized running cost is the dimensionless actuation-effort proxy
$\operatorname{mean}_{t,j}(|a_{t,j}\dot q_{t,j}|)$; since the public policy action
$a$ is not measured motor torque, this proxy has no physical energy unit. To test
whether reducing the proxy reduces mechanical work, a torque-derived evaluation
records simulator-applied joint torques $\tau$, integrates
$\sum_{t,j}|\tau_{t,j}\dot q_{t,j}|\Delta t$, and reports mean mechanical power,
work per traveled meter, cost of transport (COT, normalized by $mg$ and path
length), and elapsed task time (Sections~\ref{sec:results-main}
and~\ref{sec:results-baselines}).

\textbf{Baselines.}
(1)~\emph{Direct control}: the pretrained policy under the direct target command, the
policy being compensated and a strong baseline;
(2)~\emph{fixed scaling}: the direct command multiplied by a constant scale
($0.75$ or $0.90$), the open-loop efficiency heuristic;
(3)~\emph{reactive governor}: a measured height/tilt governor that scales the
command down when measured posture degrades.
Random and zero-command baselines are excluded as uninformative against a strong
pretrained policy. A retrieval-based compensator (VGSR) is reported in
Appendix~\ref{app:retrieval}.

\textbf{Task families.}
\emph{Harder} Go2 targets: $(2,0)$, $(1.5,1)$, $(0,1.25)$, $(-0.75,0)$, $(1,-1)$,
$(0,-1.25)$, $(-1,0.75)$, $(1.75,-0.75)$\,m: long, diagonal, lateral, and backward
goals covering the command space (the last three directions were added when the
benchmark was expanded and were never used in any tuning).
\emph{Moderate} Go2 targets: $(0.75,0)$, $(0.75,0.75)$, $(1.5,0)$, $(0,0.75)$\,m.
\emph{Holdout} Go2 directions (never used during any tuning): $(1.25,0.5)$,
$(0.5,1)$, $(-0.5,0.5)$, $(1.25,-0.5)$\,m.
ANYmal-D and H1 use the same eight harder targets as Go2.

\textbf{Protocol.}
VGCC's gate parameters (Section~\ref{sec:vgcc}) were selected on a single Go2
development seed on the harder set, then fixed for all Go2 and ANYmal-D evaluations.
Excitation collection, identification, and every gate parameter are shared
unchanged across the two quadrupeds. The humanoid retains the objective, cost
margin, response-model architecture, and correction gain, but uses more
conservative posture thresholds (eligibility, rescue, and restore fractions
$(0.93,0.90,0.96)$ and height-decline tolerance $0.8$\,cm, versus
$(0.85,0.80,0.90)$ and $2$\,cm for the quadrupeds), selected after diagnosing
humanoid posture degradation in Section~\ref{sec:results-coverage}; H1 therefore tests
morphology-adapted transfer, not zero-tuning transfer.
All reported results use disjoint evaluation seeds: harder $5$ seeds $\times$ $10$
trials $\times$ $8$ targets (400 episodes per method), moderate $3\times10$ (120),
holdout $3\times5$ (60), ANYmal-D $3\times5\times8$ (120), H1 $3\times5\times8$
(120). Per-robot response models are identified with the same protocol
(Appendix~\ref{app:impl}).
We report episode-level bootstrap 95\% CIs as descriptive uncertainty summaries.
Because episodes share a small number of environment seeds, these intervals do not
establish equivalence or noninferiority; paired seed-level differences are
reported separately, and success comparisons are phrased as ``no observed
aggregate loss'' rather than as proof of equality.

\subsection{Identifying the Command Response}
\label{sec:results-model}

Table~\ref{tab:model} reports held-out episode test metrics for the Go2 response
model.
Displacement and yaw responses are accurately predicted ($R^2 \ge 0.92$); the
effort proxy is predicted well enough to rank candidates ($R^2 = 0.75$); and
terrain-relative posture height is predicted reliably ($R^2 = 0.88$/$0.91$ for
min/mean, MAE under $9$\,mm), versus $R^2 = 0.25$ when the same model is trained on
world-frame height labels, which terrain elevation corrupts.
The excitation distribution is as consequential as the architecture: a model
trained on utility-filtered archive segments reaches $R^2=-1.9$ on lateral
displacement because curated high-quality segments underrepresent low-performance
behavior relevant to the gate.
The torque-derived mechanical-power channel used for the physical evaluations is
identified with the same protocol by a five-member ensemble and reaches
$R^2 = 0.797$ (MAE $13.0$\,W).

\textbf{Identification data efficiency.}
Re-identifying the model from episode-group subsamples degrades gracefully: with
$50\%$ of the data, displacement $R^2$ holds at $0.93$ and the effort channel drops
to $0.72$; with $25\%$ ($\approx$ one minute of parallel simulation), displacement
is still $0.92$ and the effort channel $0.62$.
Displacement, which drives the progress constraint, saturates early; accuracy on
the cost channel is what additional data improves. The $25\%$ model is evaluated in
closed loop in Appendix~\ref{app:ablation}.

\begin{table}[t]
\centering
\caption{Go2 command-response model, held-out episode test metrics ($\sim$25k excitation windows, 16-step horizon). Terrain-relative height labels are essential for the viability gate.}
\label{tab:model}
\begin{tabular}{lccc}
\toprule
Output & MAE & RMSE & $R^2$ \\
\midrule
$\Delta x$ (m)            & 0.019 & 0.025 & 0.944 \\
$\Delta y$ (m)            & 0.019 & 0.025 & 0.942 \\
$\Delta \psi$ (rad)       & 0.029 & 0.042 & 0.927 \\
effort proxy (per step)   & 0.435 & 0.625 & 0.752 \\
min.\ rel.\ height (m)    & 0.009 & 0.012 & 0.875 \\
mean rel.\ height (m)     & 0.008 & 0.010 & 0.906 \\
\midrule
min.\ height, world-frame labels & 0.039 & 0.092 & 0.247 \\
\bottomrule
\end{tabular}
\end{table}

\subsection{Compensation on Go2}
\label{sec:results-main}

The harder-target benchmark comprises eight targets evaluated over 5 seeds
$\times$ 10 trials, i.e.\ 400 episodes per method.

\begin{figure}[t]
\centering
\includegraphics[width=0.85\linewidth]{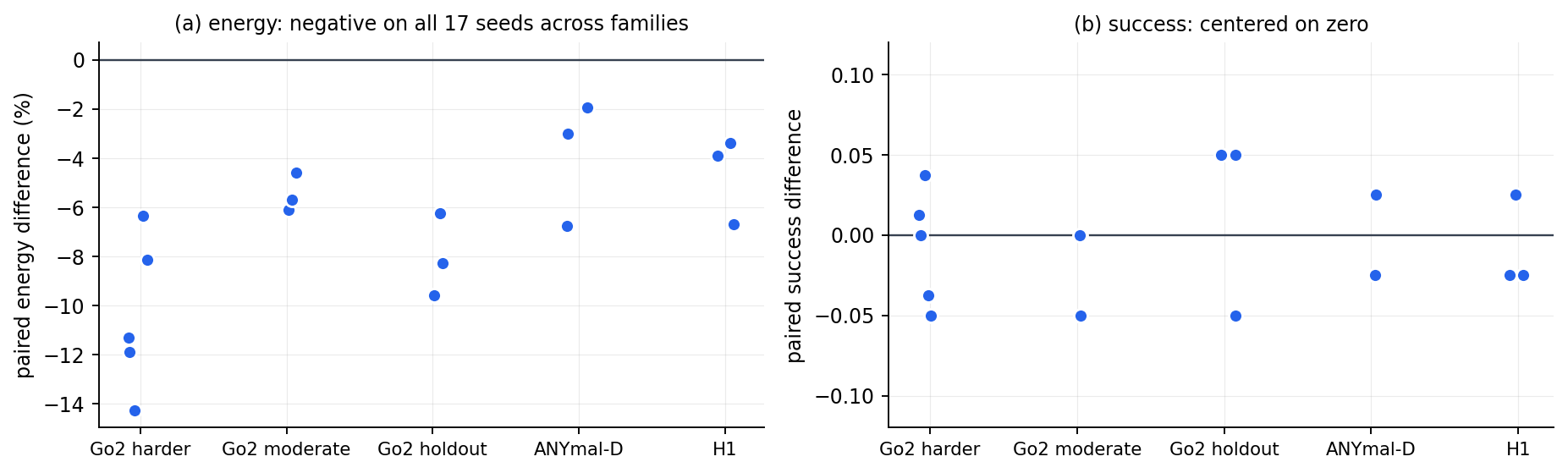}
\caption{Paired per-seed differences (VGCC $-$ direct) for all evaluation seeds. The effort-proxy difference is negative for 17/17 seeds (two-sided exact sign test $p=1.53\times10^{-5}$). Success differences straddle zero. Points are descriptive seed aggregates; the pooled sign test combines heterogeneous families.}
\label{fig:paired}
\end{figure}

\textbf{Running cost.}
VGCC reduces the actuation-effort proxy by $10.5\%$ ($1.976$ vs.\ $2.208$;
descriptive episode-bootstrap CIs $[1.938, 2.016]$ vs.\ $[2.156, 2.256]$). Observed
success is similar ($0.797$, CI $[0.760, 0.835]$, vs.\ $0.805$, CI
$[0.765, 0.843]$), as are final distance and height, with zero early terminations
for either method. The experiment was not powered as an equivalence test, so CI
overlap is not treated as evidence that success is statistically equal.
The effect is directionally consistent \emph{within} seeds: the paired per-seed
difference is negative on all five benchmark seeds ($-6.4\%$ to $-14.3\%$), and on
all 17 evaluation seeds across the Go2 families, ANYmal-D, and H1 (two-sided exact
sign test $p=1.53\times10^{-5}$), while the paired success difference ranges from
$-0.05$ to $+0.05$. Because the pooled sign test combines task families and
embodiments, it supports consistency of direction, not a common effect size.
Figure~\ref{fig:paired} visualizes both distributions.

\textbf{Torque-derived mechanical work.}
To test whether the proxy reduction reflects mechanical effort, we re-evaluate the
same configuration on three new seeds ($2$ trials $\times$ $8$ targets; 48 episodes
per method) while recording simulator-applied joint torque
(Table~\ref{tab:physical-energy}). Absolute mechanical work decreases on all three
seeds ($-8.4\%$, $-9.4\%$, $-6.6\%$; $-8.2\%$ pooled). The decomposition is
informative: VGCC lowers mean mechanical power by $16.1\%$ but takes $10.9\%$
longer, so the work saving is smaller than the power saving, and work per traveled
meter and COT fall by $6.3\%$. Observed success differs by $-2.1$ points. This
evaluation supports a simulated mechanical-work reduction for the main
configuration; it is not an electrical or hardware energy claim, and its
robustness to re-identification and to a tuned fixed scale is examined next.

\begin{table}[t]
\centering
\caption{Torque-derived mechanical-work evaluation of the main configuration on three new seeds ($n=48$ episodes per method). Work is $\sum_{t,j}|\tau_{t,j}\dot q_{t,j}|\Delta t$ using simulator-applied torque; COT normalizes by $mg$ and traveled path length. Values are descriptive because there are only three independent seeds.}
\label{tab:physical-energy}
\begin{tabular}{lrrr}
\toprule
Metric & VGCC & direct & $\Delta$ \\
\midrule
Success & 0.771 & 0.792 & $-0.021$ \\
Effort proxy & 1.933 & 2.280 & $-15.2\%$ \\
Absolute mechanical work (J) & 129.89 & 141.48 & $-8.2\%$ \\
Mean absolute power (W) & 60.29 & 71.88 & $-16.1\%$ \\
Elapsed task time (s) & 2.113 & 1.905 & $+10.9\%$ \\
Work per path length (J/m) & 111.14 & 118.60 & $-6.3\%$ \\
Cost of transport & 0.754 & 0.805 & $-6.3\%$ \\
\bottomrule
\end{tabular}
\end{table}

\subsection{Strong Baselines}
\label{sec:results-baselines}

\begin{figure}[t]
\centering
\includegraphics[width=0.72\linewidth]{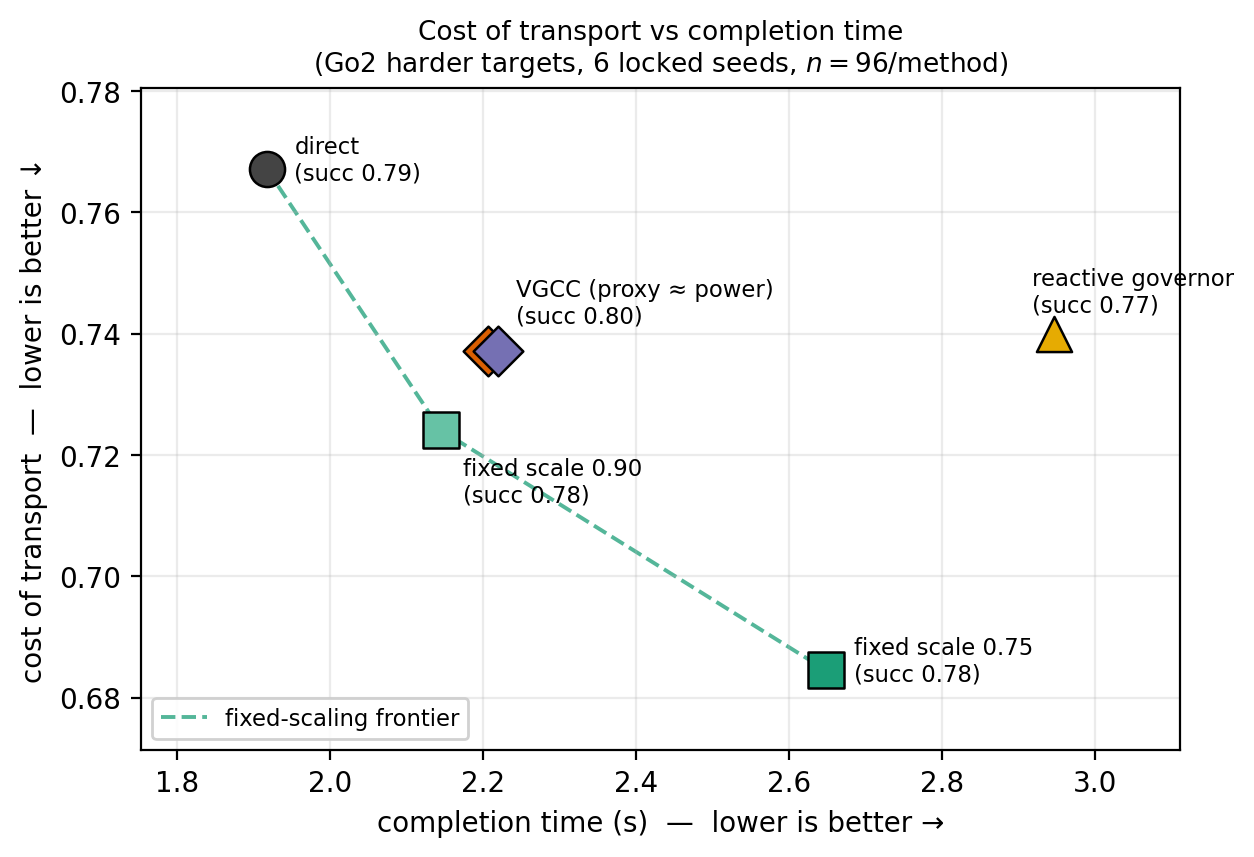}
\caption{Cost of transport versus completion time on six prespecified seeds (Go2 harder targets, $n=96$/method, identical initial states). The dashed line is the frontier defined by direct control and two fixed command scales. VGCC reduces COT relative to direct control but remains above the fixed-scaling frontier; its observed success is $0.80$. The reactive governor is dominated. Section~\ref{sec:results-headroom} evaluates whether state-conditioned allocation provides value beyond this frontier.}
\label{fig:frontier}
\end{figure}

A tuned fixed command scale is the simplest efficiency heuristic and defines the
reference frontier against which learned adapters should be evaluated. We compare against it, a
reactive governor, and direct control on a prespecified six-seed evaluation with
torque-derived power, all from identical initial states. VGCC is shown under both
cost channels so the result does not depend on which channel is optimized
(Table~\ref{tab:objective-audit}, Figure~\ref{fig:frontier}). Fixed $0.90$ provides
the closest completion-time comparison; fixed $0.75$ represents the slower,
lower-COT end of the frontier.

VGCC reduces mechanical work ($5.1\%$), power ($16\%$), and COT ($3.9\%$) below
direct control, with the highest success of any method ($0.802$ vs.\ $0.792$).
It does \emph{not} dominate the fixed-scaling frontier: at matched time, fixed
$0.90$ reaches lower COT ($0.724$ vs.\ $0.737$), and fixed $0.75$ the lowest COT
($0.685$) at a $38\%$ time penalty. VGCC sits just above this frontier
(Figure~\ref{fig:frontier}).

A higher-capacity baseline yields the same ordering. An 81-sequence sampling MPC plans four
high-level steps over $\{0.75,0.90,1.0\}$ with a separately trained recursive
macro-transition ensemble and task-value model (Appendix~\ref{app:impl}). On three
of the six seeds it reaches COT $0.757$---below direct ($0.765$) but above VGCC
($0.736$) and fixed $0.90$ ($0.721$). Thus, VGCC does not outperform the tuned
fixed scale. The remaining questions concern transfer of the same configuration
(Section~\ref{sec:results-coverage}) and whether state-conditioned allocation can
justify adapter complexity over this frontier
(Section~\ref{sec:results-headroom}).

\textbf{Contribution of reduced speed.} COT and work per meter account for
distance traveled but are not generally invariant to speed. Removing the progress
floor increases the proxy reduction from $12.2\%$ to $19.6\%$ while reducing
success by $3.7$ percentage points (Appendix~\ref{app:ablation}). VGCC increases
episode duration by $17\%$, compared with $40\%$ for fixed $0.75$.

\begin{table}[t]
\centering
\caption{Prespecified frontier comparison on six seeds ($n=96$/method), ordered by completion time. All methods are evaluated from \emph{identical} initial states (paired resets), so differences are within condition. VGCC is shown under both cost channels. Bold denotes the best value in each column. VGCC reduces work, power, and COT relative to direct control and attains the highest observed success, whereas the tuned fixed scales define a lower COT--time frontier.}
\label{tab:objective-audit}
\begin{tabular}{lrrrrr}
\toprule
Method & Success & Work (J) $\downarrow$ & Power (W) $\downarrow$ & Time (s) $\downarrow$ & COT $\downarrow$ \\
\midrule
direct & 0.792 & 135.07 & 68.09 & \textbf{1.918} & 0.767 \\
fixed scaling ($0.90$) & 0.781 & 126.85 & 57.28 & 2.145 & 0.724 \\
VGCC (proxy cost) & \textbf{0.802} & 128.20 & 56.99 & 2.207 & 0.737 \\
VGCC (power cost) & \textbf{0.802} & 128.28 & 56.67 & 2.220 & 0.737 \\
fixed scaling ($0.75$) & 0.781 & \textbf{119.42} & \textbf{44.05} & 2.648 & \textbf{0.685} \\
reactive governor & 0.771 & 128.51 & 44.69 & 2.947 & 0.740 \\
\bottomrule
\end{tabular}
\end{table}

\subsection{Coverage and Transfer}
\label{sec:results-coverage}

Before auditing recoverability, we assess whether the VGCC results are specific
to the Go2 harder-target family. Table~\ref{tab:coverage} evaluates the fixed configuration
on remaining Go2 target families and two further robots, without per-family or
per-robot objective tuning.

\begin{table}[t]
\centering
\caption{Coverage across task families and embodiments. $\Delta$ is the relative change in the actuation-effort proxy. Success CIs are descriptive, not equivalence tests. Go2 and ANYmal-D use the frozen quadruped configuration; H1 uses morphology-adapted posture thresholds (Section~\ref{sec:setup}).}
\label{tab:coverage}
\begin{tabular}{llcccc}
\toprule
Family ($n$/method) & Method & Success $\uparrow$ & Effort $\downarrow$ & Effort $\Delta$ & Final height (m) \\
\midrule
\multirow{2}{*}{Go2 moderate (120)}
 & VGCC   & 0.975 {\scriptsize$[0.942, 1.000]$} & \textbf{1.743} {\scriptsize$[1.689, 1.802]$} & $-5.4\%$ & 0.356 \\
 & direct & \textbf{0.992} {\scriptsize$[0.975, 1.000]$} & 1.844 {\scriptsize$[1.763, 1.932]$} & N/A & \textbf{0.362} \\
\midrule
\multirow{2}{*}{Go2 holdout (60)}
 & VGCC   & \textbf{0.933} {\scriptsize$[0.867, 0.983]$} & \textbf{1.936} {\scriptsize$[1.836, 2.037]$} & $-8.0\%$ & \textbf{0.334} \\
 & direct & 0.917 {\scriptsize$[0.833, 0.983]$} & 2.105 {\scriptsize$[1.982, 2.234]$} & N/A & 0.333 \\
\midrule
\multirow{2}{*}{ANYmal-D, 8 targets (120)}
 & VGCC   & \textbf{0.883} {\scriptsize$[0.825, 0.942]$} & \textbf{1.367} {\scriptsize$[1.317, 1.422]$} & $-3.9\%$ & 0.500 \\
 & direct & 0.875 {\scriptsize$[0.817, 0.933]$} & 1.422 {\scriptsize$[1.373, 1.471]$} & N/A & 0.500 \\
\midrule
\multirow{2}{*}{H1, 8 targets (120)}
 & VGCC   & 0.967 {\scriptsize$[0.933, 0.992]$} & \textbf{0.690} {\scriptsize$[0.665, 0.716]$} & $-4.7\%$ & 0.928 \\
 & direct & \textbf{0.975} {\scriptsize$[0.942, 1.000]$} & 0.724 {\scriptsize$[0.696, 0.752]$} & N/A & \textbf{0.941} \\
\bottomrule
\end{tabular}
\end{table}

\textbf{Moderate and holdout targets.}
The effort proxy drops on both families ($-5.4\%$ and $-8.0\%$), while observed
success differs by $-1.7$ and $+1.6$ percentage points. The holdout family is the
most informative case: these directions were never observed during any tuning, so
the frozen gate generalizes across the command space, a property the retrieval
variant does not possess (Appendix~\ref{app:retrieval}: the archive variant
\emph{increases} the proxy on these directions).

\textbf{Third embodiment without parameter retuning: ANYmal-D.}
Applying the same protocol to a different quadruped reduces the effort proxy by
$3.9\%$; observed success is $0.883$ vs.\ $0.875$, with identical final posture and
zero terminations. The paired effort difference is negative on all three ANYmal
seeds. Since no gate or objective parameter was adjusted, this is consistent with a
cross-quadruped effect rather than a Go2-specific result. A single robot and three
seeds are nevertheless insufficient to establish broad embodiment generality.

\textbf{H1 humanoid: posture floors are morphology-dependent.}
Transfer is not zero-shot. With the quadruped posture thresholds, VGCC reduces H1
posture substantially on long forward and diagonal targets (final height
$0.47$\,m for the $2$\,m target). The taller center of mass and smaller support
region require more conservative posture constraints. Tightening eligibility, rescue, and
restore thresholds---leaving the objective, cost margin, architecture, and
correction gain unchanged---recovers a $-4.7\%$ effort-proxy reduction with
success $0.967$ vs.\ $0.975$ and negative paired differences on all three seeds.
Because thresholds were changed after observing the initial posture degradation, this is
morphology-adapted evidence, not zero-shot transfer.

\subsection{Ablations}
\label{sec:results-ablation}

A paired component study (Appendix~\ref{app:ablation}, $n=80$ per variant) isolates
the role of each mechanism. Removing the posture gate forfeits most of the proxy
saving ($-4.8\%$ vs.\ $-12.2\%$) and reduces success; removing the progress floor
deepens the saving to $-19.6\%$ but reduces success by $3.7$ points; and full
command substitution ($\alpha=1$) has the largest success cost ($-5.0$ points),
which is why the correction is bounded. The gate keeps predicted savings inside the
model's well-behaved posture regime, the progress floor limits behavioral drift,
and bounded blending avoids the success cost of substitution. Quarter-data
identification retains displacement $R^2=0.92$ but halves the realized proxy
reduction, tracking the cost-channel degradation in Section~\ref{sec:results-model}.
These results characterize the mechanisms underlying VGCC's improvement relative
to direct control; they do not establish state-allocation value. The subsequent
audit evaluates that distinction.

\subsection{Adapter Necessity Audit}
\label{sec:results-headroom}

\begin{figure}[t]
\centering
\includegraphics[width=\linewidth]{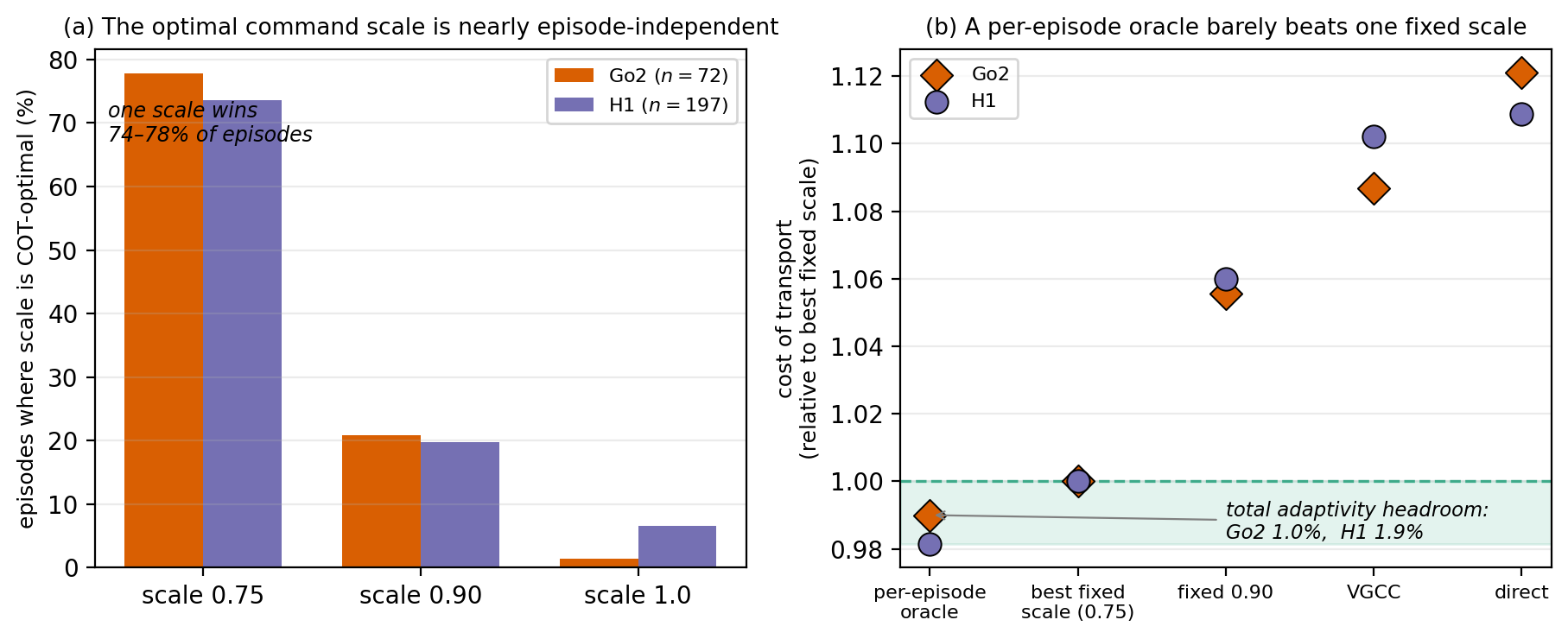}
\caption{Stage 1 of the adapter necessity audit, on paired-successful episodes (identical initial states). \textbf{(a)} The COT-optimal command scale is nearly the same in every episode: a single scale is optimal in $74$--$78\%$ of episodes on both robots. \textbf{(b)} Cost of transport relative to the best single fixed scale ($=1$): a finite-grid per-episode oracle lies only $1.0\%$ (Go2) and $1.9\%$ (H1) below that fixed scale. This is an episode-level diagnostic over three scales, not an upper bound on within-episode switching or commands outside the grid.}
\label{fig:finding}
\end{figure}

Controller-level comparisons do not distinguish limited state-dependent
opportunity from an inability of the learned representation to recover that
opportunity. We therefore audit adapter necessity in three stages. Stage 1
measures opportunity at episode and same-state resolutions. Stage 2 cross-fits a
treatment-effect selector and compares it with both a training-selected fixed
reference and a randomized mixture with identical marginal command frequencies.
Stage 3 propagates source-cluster and learner-fitting uncertainty into the formal
decision. Improvement over the matched mixture attributes value to conditioning;
improvement over the fixed reference establishes deployment value.

For every episode, we observe the cost of transport of the direct command
($1.0$), fixed $0.90$, and fixed $0.75$ from identical initial states. We compare the
best single \emph{global} scale (minimizing pooled COT) against a finite-grid oracle that
picks the best scale \emph{per episode}. The oracle upper-bounds what any selector
that holds one scale from this set for the whole episode could extract; the gap
between it and the best global scale quantifies episode-level adaptivity
headroom. A per-step switch (VGCC) or plan (the sampling MPC) is not formally
bounded by an episode-level oracle. In the present evaluation, however, neither
outperforms the best fixed scale in Table~\ref{tab:objective-audit}. Corrections
outside the scale family are not searched by the oracle; VGCC's candidate set does
include redirected and re-timed variants, and their availability does not change
its position against the fixed scales.

The episode-level gap is small (Figure~\ref{fig:finding}): the per-episode oracle
improves on the best single scale by only $1.0\%$ on Go2 ($n=72$ fully successful
paired episodes; episode-bootstrap $95\%$ CI $[0.5, 1.7]\%$) and $1.9\%$ on H1
($n=197$; CI $[1.1, 3.0]\%$). H1 here is a descriptive second-embodiment Stage~1
screen with the same three-scale grid: the same near-constant global slowdown
pattern. It is not a cluster-level confirmation. The per-episode optima clarify
this pattern: scale $0.75$ is selected in $78\%$ of
Go2 episodes and $74\%$ of H1 episodes, and COT decreases monotonically over the
evaluated scales. Thus the efficiency-optimal command within this grid is nearly
constant across episodes. This result alone does \emph{not} bound
within-episode switching, redirected commands, or a richer command set. It is a
preliminary diagnostic: when deployment permits only one scale per episode, the
observed headroom provides limited motivation for learning a selector over this
grid.

\textbf{Stage 1: same-state headroom.}
We next isolate a single intervention from long-horizon controller effects.
At 480 query states from ten source seeds, each candidate prefix
$s\in\{0.75,0.90,1.00\}$ is replayed from the \emph{same} observation and then
followed by the same direct-control continuation. Replay is exact after freezing
the terrain curriculum: the query-observation reconstruction error is zero.
A candidate is eligible only if its realized success is no worse than direct
and its remaining completion time is at most $1.10$ times direct. The
same-state oracle then chooses the eligible prefix with minimum total mechanical
work. Unlike the episode-level analysis, this test shows meaningful local
variation: the oracle reduces work by $5.21\%$ while slightly increasing
success ($0.698$ versus $0.681$).

This reported action set contains only constant scale prefixes. It aligns with
the primitive action family used inside the sampling MPC, although it does not
audit arbitrary four-step scale sequences. It does not align with VGCC's full
candidate set, which also contains heading- and yaw-rate interventions.
Consequently, the result characterizes scale personalization only. Evaluation of
the opportunity available to VGCC requires exact replay of at least one
prespecified heading-offset family and one yaw-rate-offset family under the same
continuation protocol.

\textbf{Stage 2: recoverability and attribution.}
Existence is not predictability. We train a five-member treatment-effect ensemble
to predict each candidate's work and time \emph{relative to direct}, using the
deployment observation, target, and identified macro-response means and
uncertainties. Every reported prediction is leave-one-source-seed-out. A
conservative selector intervenes only when the work upper bound is negative,
the time upper bound satisfies the budget, and the eligibility lower bound
exceeds $0.5$. To separate state information from action frequency, its matched
randomized mixture draws scales independently of state using the selector's
own scale frequencies within each held-out fold.

\begin{table}[t]
\centering
\caption{Descriptive scale-prefix analysis on ten exact-replay source seeds (480 same-state queries, scales $\{0.75,0.90,1.00\}$, success no worse than direct, time $\le1.10\times$ direct). Work changes are relative to the comparator named in each row; negative values denote lower work. The matched-mixture row isolates the part attributable to state--action matching.}
\label{tab:adapter-audit}
\begin{tabular}{lccc}
\toprule
Audit quantity & Work vs.\ direct & Constraint violations & Interpretation \\
\midrule
same-state oracle & $-5.21\%$ & $0\%$ by construction & available local headroom \\
LOSO learned selector & $-0.68\%$ & $2.08\%$ & total selected gain \\
matched randomized mixture & $-0.13\%$ & $3.93\%$ & action-frequency gain \\
selector vs.\ matched mixture & $-0.55\%$ & --- & state-allocation component \\
\bottomrule
\end{tabular}
\end{table}

The learned selector activates on $13.3\%$ of queries and matches the oracle
choice on $46.3\%$. Its candidate relative-work MAE is $11.6\%$, versus only
$5.2\%$ total oracle headroom and $0.55\%$ recovered state-allocation gain.
It beats the matched mixture on work in seven of ten held-out seeds, but the
effect remains below one percent. With source-seed cluster bootstrap,
$H_{\rm alloc}=0.55\%$ has a descriptive $95\%$ CI of $[0.14,1.03]\%$, and
$q=2.08\%$ has CI $[1.04,3.33]\%$. These intervals resample fitted LOSO fold
summaries and therefore do not include learner-refitting uncertainty. Moreover,
the archived analysis did not evaluate $H_{\rm dep}$ against a cross-fitted fixed
reference. Under Eq.~\ref{eq:audit-decision}, it consequently receives no formal
decision. An observation-only selector on the same ten clusters has
$H_{\rm alloc}=0.41\%$ with CI $[0.16,0.68]\%$, and adding the recursive macro predictions
raises the point estimate only from $0.41\%$ to $0.55\%$. The five-seed feature
ablation in Appendix~\ref{app:feature-ablation} reaches the same qualitative
conclusion. Identification is therefore tested as an optional source of
prospective decision features, rather than credited with a bound that a
counterfactual sweep computes without it.

\textbf{Stage 3: action-aligned confirmatory audit.}
We collect twenty independent source clusters for each Go2 query distribution
induced by direct control, evaluated VGCC, and evaluated MPC. Each cluster
contains 48 queries, for 960 queries per distribution. At every query, exact
branch replay evaluates
$\{\text{direct},0.75,0.90,\Delta\psi=\pm0.3,
\Delta\omega_z=\pm0.2\}$ after the deployed bounded blend
($\alpha=0.5$) for four macro steps. The learner uses observation-only features,
the fixed reference is selected from training clusters only, and 200
source-cluster bootstrap replicates refit the complete learner and evaluate it
on out-of-bootstrap clusters. We use the prespecified headline thresholds
$\delta_{\rm dep}=\delta_{\rm alloc}=1\%$ and $\kappa=5\%$.

\begin{table}[t]
\centering
\caption{Action-aligned confirmatory audit. Each row contains twenty independent
source clusters, 960 queries, and 200 full learner refits. Gains are positive
when the learned selector reduces work. Values are means with cluster-refit
95\% intervals; $q$ is the violation rate. Decisions use
$(\delta_{\rm dep},\delta_{\rm alloc},\kappa)=(1\%,1\%,5\%)$.}
\label{tab:confirmatory-audit}
\begin{tabular}{lcccc}
\toprule
Embodiment/query distribution & $H_{\rm dep}$ (\%) & $H_{\rm alloc}$ (\%) & $q$ (\%) & Decision \\
\midrule
Go2/direct & $0.60\ [-0.18,1.22]$ & $0.53\ [-0.30,0.78]$ & $2.08\ [0.83,5.49]$ & \textsc{No-Go} \\
Go2/VGCC & $1.34\ [0.63,2.03]$ & $0.82\ [0.09,1.28]$ & $3.96\ [1.78,6.25]$ & \textsc{Abstain} \\
Go2/MPC & $0.70\ [-0.08,1.73]$ & $0.64\ [-0.07,1.20]$ & $3.33\ [1.19,5.73]$ & \textsc{Abstain} \\
H1/direct & $-0.35\ [-0.81,0.71]$ & $0.12\ [-0.19,0.67]$ & $1.67\ [0.00,2.68]$ & \textsc{No-Go} \\
\bottomrule
\end{tabular}
\end{table}

No real-domain row returns \textsc{Go}. Go2/direct returns \textsc{No-Go}
because the upper allocation bound, $0.78\%$, is below the required $1\%$;
its positive point estimates therefore do not establish meaningful
state-dependent personalization. VGCC has the strongest mean deployment gain,
$1.34\%$, but its deployment lower bound is $0.63\%$, its allocation lower bound
is $0.09\%$, and its violation upper bound is $6.25\%$. It consequently returns
\textsc{Abstain}, not a near-\textsc{Go}. MPC is likewise unresolved on all
three boundaries.

\textbf{Why both comparators matter.}
The comparator pair changes the scientific conclusion, not only the reported
effect size. On VGCC queries, the selector improves by $1.34\%$ over direct
control (also the cross-fitted fixed reference), but only $0.82\%$ remains when
its marginal action frequencies are held fixed. On H1, comparison with direct
control suggests a $0.16\%$ gain and the matched-mixture attribution is
$0.12\%$; nevertheless, the selector is $0.35\%$ worse than the cross-fitted
scale-$0.90$ reference. A direct-only comparison would obscure the fixed
operating point, whereas a fixed-only comparison would not determine whether
any remaining gain came from context-dependent allocation. Requiring both
therefore prevents two distinct false recommendations.

\textbf{Decision-resolution diagnostic.}
We additionally perform a post-hoc planning calculation, without changing the
formal decisions or thresholds. If each directional confidence-bound width
shrinks as $N^{-1/2}$ while the point estimate remains fixed, VGCC would require
approximately 86 clusters for the deployment lower bound and 97 for the
violation upper bound to cross their respective boundaries. Its mean allocation
gain, however, is $0.82\%$, below the $1\%$ threshold; variance reduction alone
cannot place the allocation lower bound above $1\%$ at that point estimate.
MPC's mean deployment and allocation gains are likewise both below $1\%$.
Consequently, collecting more clusters alone is not a credible route to a
\textsc{Go}; it can resolve uncertainty, but a positive recommendation would
also require a more recoverable representation, intervention family, or
deployment domain. Appendix~\ref{app:audit-resolution} gives the calculation.

\textbf{H1 query-distribution correction.}
The earlier four-cluster pilot used randomly initialized, dimension-matched
surrogates to generate visited states and remains excluded. The confirmatory H1
collection instead samples all queries under direct deployment control, without
either surrogate, and evaluates the same seven action-aligned interventions on
twenty independent clusters (960 queries). It returns \textsc{No-Go} because
the deployment-gain upper bound, $0.71\%$, is below $\delta_{\rm dep}$. Thus the
second-embodiment result now targets a deployment-relevant estimand rather than
only validating branch reconstruction.

\begin{figure}[t]
\centering
\includegraphics[width=\linewidth]{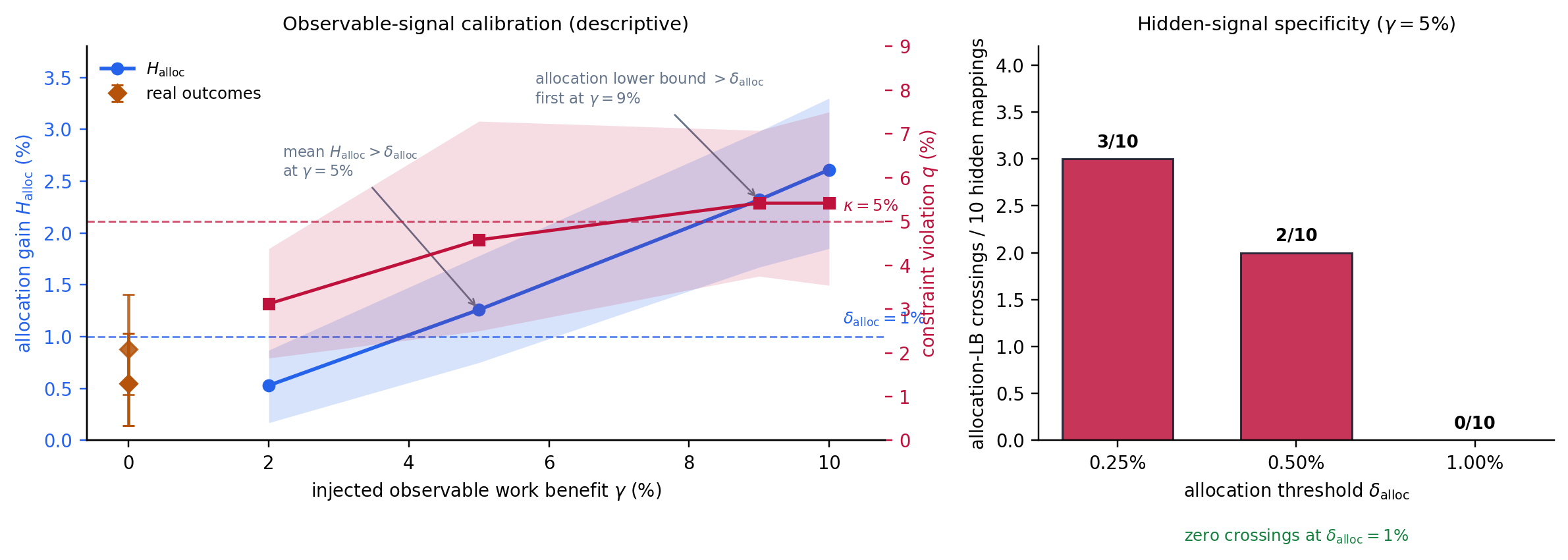}
\caption{Descriptive outcome-level calibration on ten source clusters.
\textbf{Left:} observable-signal sweep for allocation gain and violation rate.
\textbf{Right:} ten hidden-signal mappings characterize spurious allocation
findings at thresholds from $0.25\%$ to $1\%$. These archived curves do not
include deployment gain or learner-refit uncertainty and therefore do not yield
formal decisions under Eq.~\ref{eq:audit-decision}.}
\label{fig:audit-calibration}
\end{figure}

\textbf{Calibration controls.}
We assess sensitivity and specificity with semi-synthetic controls that retain
all 480 real Go2 query states, outcomes, and cross-seed splits
(Figure~\ref{fig:audit-calibration};
Table~\ref{tab:audit-calibration}). These archived controls characterize the
allocation and violation channels; because they lack $H_{\rm dep}$ and learner
refitting, they are not evaluations of the revised formal rule and are not a
second robot domain. For the
\emph{observable-signal} control, scale $0.90$ receives a fractional work bonus
on nonnegative-lateral targets and scale $0.75$ otherwise; the pre-intervention
target exposes which scale is preferred. We sweep the bonus $\gamma$ from $2\%$
to $10\%$. A \emph{hidden-signal} control assigns the preferred scale per query
and omits it from all features, creating oracle headroom no selector should
systematically recover.

\begin{table}[t]
\centering
\caption{Archived allocation/safety calibration on ten source clusters.
Intervals resample fitted fold summaries and are descriptive; the table does not
report formal three-channel decisions.}
\label{tab:audit-calibration}
\begin{tabular}{lcc}
\toprule
Condition & $H_{\rm alloc}$, 95\% CI & $q$, 95\% CI \\
\midrule
real outcomes & $0.55\ [0.14,1.03]\%$ & $2.08\ [1.04,3.33]\%$ \\
observable signal, $2\%$ & $0.53\ [0.17,0.87]\%$ & $3.12\ [1.88,4.38]\%$ \\
observable signal, $5\%$ & $1.26\ [0.75,1.78]\%$ & $4.58\ [2.50,7.29]\%$ \\
observable signal, $9\%$ & $2.32\ [1.67,2.98]\%$ & $5.42\ [3.75,7.08]\%$ \\
observable signal, $10\%$ & $2.61\ [1.85,3.30]\%$ & $5.42\ [3.54,7.50]\%$ \\
\bottomrule
\end{tabular}
\end{table}

Mean $H_{\rm alloc}$ exceeds $1\%$ once $\gamma=5\%$, while stronger bonuses
also raise intervention and violation rates. Ten independent $5\%$
hidden-signal mappings exceed the allocation-only lower-bound thresholds in
three of ten cases at $0.25\%$, two at $0.5\%$, and zero at $1\%$. Thus $1\%$
is retained as a conservative allocation threshold for the present calibration,
not as a universal constant. A deployment domain must additionally prespecify
$\delta_{\rm dep}$ and calibrate all three channels. Full curves are in
Appendix~\ref{app:audit-controls}.

The learner-level positive control contains twenty synthetic source clusters.
Observed target sign identifies which non-direct action receives a $20\%$ work
benefit, and both candidates satisfy the success and time constraints. The
control executes the complete analysis protocol: treatment-effect learning,
conservative selection, both comparators, cluster-level refitting, and the
decision rule. With $\delta_{\rm dep}=\delta_{\rm alloc}=1\%$ and $\kappa=5\%$,
the three-channel rule returns \textsc{Go}. This establishes procedural
reachability. Alongside the real-domain decisions in
Table~\ref{tab:confirmatory-audit}, it shows that the absence of a real
\textsc{Go} is not caused by an unreachable decision rule.

\textbf{Terrain heterogeneity and the efficiency frontier.}
If rough terrain benefits more from command attenuation than flat terrain, a
terrain-conditioned scale could outperform a single fixed scale. We evaluate
this hypothesis with flat and uniformly rough ($2$--$10$\,cm)
variants, a re-identified response model, and the scale grid extended to $0.60$
(Appendix~\ref{app:hetero}, Table~\ref{tab:hetero}). COT decreases monotonically
with scale on \emph{both} terrains: roughness raises the cost level but has
little effect on the location of its minimum. A shared completion-time budget
that restricts \emph{where} attenuation can be applied yields $1.6\%$ of total work
(bootstrap $95\%$ CI $[0.8, 8.4]$) at the tightest budgets, and nothing once
attenuation is permitted everywhere. Thus roughness does not create a distinct
operating-point optimum in this evaluation.

Episode-level estimates are specific to the evaluated grid. The archived
exact-replay analysis covers Go2 scale prefixes on ten descriptive source
clusters. The confirmatory audit adds direct, scale, heading, and yaw
interventions on three Go2 query distributions and deployment-representative H1
queries, but evaluates only one heading-offset and one yaw-offset magnitude in
each direction rather than every candidate magnitude or multi-step sequence.
Unevaluated terrain properties (slopes, compliance, and friction) could reshape
rather than merely shift the cost curve; richer interfaces such as gait
parameters~\citep{margolis2022walk} are where tracking may not flatten structure.
Accordingly, the findings establish that adaptivity headroom is measurable and
that recoverable allocation gain is insufficiently large or precise for the
evaluated benchmark family; they do not imply the absence of gain under broader
interventions or domains.

\section{Limitations and Future Work}
\label{sec:limitations}

\textbf{Energy validity.}
Large-sample results optimize a dimensionless actuation-effort proxy; torque
audits support lower simulated mechanical work and COT versus direct control, but
not versus the fixed-scaling frontier. Absolute joint work is not electrical
energy; hardware power measurement remains necessary for a real energy claim.

\textbf{Temporal confounding.}
VGCC episodes run $10$--$14\%$ longer despite the progress floor. An explicit
completion-time budget would make residual savings unambiguously iso-time; the
mixed-terrain shared-budget estimate is only $1.6\%$ of work with a wide CI
$[0.8, 8.4]$.

\textbf{Statistics and action-space alignment.}
Episode bootstrap intervals ignore nesting in seeds; overlapping success CIs are
not equivalence tests. The archived scale-only Go2 analysis resamples fitted
LOSO summaries from ten source clusters and remains descriptive. The
confirmatory analysis uses twenty independent source clusters per distribution
and 200 full learner refits, but finite bootstrap Monte Carlo error remains.
Its direct, scale, heading, and yaw interventions align with the principal
action families of VGCC, while sampling only a sparse subset of magnitudes and
no multi-step redirected-command sequences. It therefore supports inference
about the tested action families and query distributions, not exhaustive
coverage of VGCC's approximately 70 candidates. Thresholds, cluster counts, and
action families should be preregistered in subsequent domains.

\textbf{Recoverability and inferential scope.}
Local scale-prefix oracle headroom of $5.2\%$ with recovered allocation gain of
$0.55\%$ is evidence about the present representation and data, not a theorem
that no learner can recover the oracle. Semi-synthetic controls
(Figure~\ref{fig:audit-calibration}) probe allocation sensitivity and
specificity; a twenty-cluster learner-level synthetic control returns
three-channel \textsc{Go}. The real action-aligned analysis resolves Go2/direct
and H1/direct as \textsc{No-Go}, while VGCC and MPC remain
\textsc{Abstain}. Further evaluation requires hardware energy measurement,
additional clusters to resolve the boundary channels, and a better selector or
domain to raise allocation value rather than merely narrow its interval.
Broader heading/yaw magnitudes and command sequences remain untested.

\textbf{Scope.}
All results are simulated target reaching with velocity-commanded policies on
three embodiments. The framework assumes a command interface with redundancy and
fast closed-loop response. Untested: end-effector-commanded manipulation,
style-commanded character control, hardware transfer, and richer gait interfaces
where the tracking objective may not flatten state-dependent structure.

\section{Conclusion}
\label{sec:conclusion}

We present an adapter necessity audit for determining whether a frozen
locomotion interface supports a learned command adapter. The audit separates
global operating-point gain, same-state counterfactual headroom, recoverable
state-allocation gain over a frequency-matched mixture, and deployment gain over
a cross-fitted fixed reference. Its \textsc{Go}/\textsc{No-Go}/\textsc{Abstain}
rule jointly evaluates deployment value, attribution value, and violations with
learner-refit source-cluster uncertainty. Closed-loop identification
provides optional features and is accurate on average, but average accuracy does
not imply treatment-effect ranking: on Go2, $5.2\%$ local scale-prefix headroom
yields only $0.55\%$ descriptive allocation gain. In the action-aligned
twenty-cluster audit, Go2/direct and H1/direct return \textsc{No-Go}, whereas
VGCC and MPC return \textsc{Abstain}; no real-domain distribution returns
\textsc{Go}. A twenty-cluster learner-level synthetic control does return
three-channel \textsc{Go}, validating procedural reachability. These findings
cover representative direct, scale, heading, and yaw interventions, but not
VGCC's complete candidate grid or hardware energy. Priority extensions are
hardware power measurement and a deployment domain with recoverable
personalization. For the unresolved Go2 distributions, more clusters can narrow
the decision bounds but cannot create a \textsc{Go} while the allocation point
estimates remain below the practical-value threshold.

\subsubsection*{Reproducibility}
All experiments use public Isaac Lab environments and public pretrained RSL-RL
checkpoints. The project repository provides the experimental protocols,
analysis procedures, and aggregate results used in this study.

\bibliographystyle{unsrtnat}
\bibliography{references}

\clearpage
\appendix
\section{Experimental and Model Details}
\label{app:impl}

This section specifies the data collection, model training, controller
configuration, and counterfactual evaluation protocols used in the experiments.

\textbf{Excitation and dataset.}
Excitation overrides the converged policy's command with commands drawn uniformly
over the admissible box $[-1,1]^3$ in $(v_x, v_y, \omega_z)$, resampled every 25
low-level steps; with probability $0.3$ the drawn command is scaled by $0.3$,
oversampling the low-magnitude regime that proportional control visits near the
goal. Collection runs 64 parallel environments (8{,}000 steps per environment for
H1). Constant-command windows of 16 low-level steps ($0.32$\,s at the 50\,Hz
control rate) are sliced at stride 4 within episodes, giving the $\approx$25k Go2
windows of Table~\ref{tab:model}. Terrain-relative height labels are recovered
from the policy's own height-scan observation (mean scan return plus its
$0.5$\,m sensor offset).

\textbf{Response model and training.}
$f_\phi$ is a multilayer perceptron with two hidden layers of width 256 (ReLU,
dropout $0.1$), mapping the full policy observation, with the command slice
replaced by the candidate command, to the six response targets; inputs and
targets are standardized. Training uses AdamW (learning rate $10^{-3}$, weight
decay $10^{-4}$), batch size 1{,}024, 200 epochs, keeping the checkpoint with the
best held-out loss. The held-out set is $20\%$ of episode groups, so no episode
straddles the split; Table~\ref{tab:model} reports this held-out set. The
torque-derived mechanical-power channel is identified with the same protocol by a
five-member ensemble.

\textbf{Candidate set.}
The structured set $\mathcal{C}$ around the direct command comprises: uniform
scalings $\{0.4, 0.6, 0.8, 1.2\}\,c_{\text{goal}}$ and $c_{\text{goal}}$ itself;
yaw-rate offsets of $\{\pm0.2, \pm0.4\}$\,rad/s applied to $c_{\text{goal}}$; and
a polar grid of absolute speeds $\{0.25, 0.5, 0.75, 1.0\}$\,m/s at heading
offsets $\{0, \pm0.3, \pm0.6\}$\,rad around the target bearing with yaw-rate
scalings $\{0, 0.5, 1.0\}$ of the direct yaw command. All candidates are clipped
to the admissible box $[-1,1]^3$ and deduplicated, leaving $\approx$70.

\textbf{Controller constants.}
One high-level decision is taken every $E=4$ low-level steps. The gate scalars
shared by all three robots are: correction gain $\alpha=0.5$ with anneal
distance $d_{\text{anneal}}=0.5$\,m (the correction fades out linearly inside
$0.5$\,m of the target), progress floor $\beta=0.9$, and cost margin
$\epsilon=0.1$; the cost channel $\hat C$ is the effort proxy for the main
configuration and predicted mechanical power for the power configuration of
Table~\ref{tab:objective-audit}. The morphology-dependent posture floors are the
eligibility, rescue, and restore fractions of nominal height with the
height-decline tolerance $\delta$: $(0.85, 0.80, 0.90)$ with $\delta=2$\,cm for
both quadrupeds, $(0.93, 0.90, 0.96)$ with $\delta=0.8$\,cm for H1
(Section~\ref{sec:setup}).

\textbf{VGCC decision rule.}
Algorithm~\ref{alg:vgcc} specifies one high-level decision. If no candidate is
eligible or none satisfies the cost margin, the controller executes
$c_{\text{goal}}$.

\begin{algorithm}[t]
\caption{Viability-Gated Command Compensation (one high-level step)}
\label{alg:vgcc}
\begin{algorithmic}[1]
\Require frozen policy $\pi_\theta$, response model $f_\phi$, target $g$, correction gain $\alpha$, anneal distance $d_{\text{anneal}}$, and fixed scalars $(\beta,h_{\rm floor},h_{\rm resc},\delta,\epsilon)$
\State Compute body-frame target $g$, direct command $c_{\text{goal}}$ (Eq.~\ref{eq:cgoal}), and the candidate set $\mathcal{C}$ around $c_{\text{goal}}$
\State Batch-predict $f_\phi(o,c)$ for all $c\in\mathcal{C}$; set $\alpha_t=\alpha\operatorname{clip}(d/d_{\text{anneal}},0,1)$
\If{measured posture is below the operating band}
  \State $c\gets c_{\text{goal}}$ \Comment{restore}
\ElsIf{direct control is predicted to breach the posture floor $h_{\rm resc}$}
  \State $c^\ast\gets\arg\max_{\mathcal{C}:\,\hat G\ge 0}\hat h_{\min}$;\quad $c\gets c_{\text{goal}}+\alpha_t(c^\ast-c_{\text{goal}})$ \Comment{rescue}
\Else
  \State $c^\ast\gets\arg\min_{\mathcal{C}\,\text{eligible (Eq.~\ref{eq:vgc-eligible})}}\hat C$
  \If{$\hat C(c^\ast)<(1-\epsilon)\,\hat C(c_{\text{goal}})$ (Eq.~\ref{eq:vgc-cost})}
    \State $c\gets c_{\text{goal}}+\alpha_t(c^\ast-c_{\text{goal}})$ \Comment{compensate}
  \Else
    \State $c\gets c_{\text{goal}}$ \Comment{exact direct fallback}
  \EndIf
\EndIf
\State Execute $\pi_\theta$ with $c$ for $E$ low-level steps
\end{algorithmic}
\end{algorithm}

\textbf{Sampling MPC baseline.}
The MPC of Section~\ref{sec:results-baselines} does \emph{not} recursively use
the six-output response model $f_\phi$: that model has no next-observation
output and cannot be rolled forward. Instead, MPC uses a separately trained
five-member macro-transition ensemble
\[
F_\eta(o_t,c_t)\mapsto
(\widehat{o_{t+4}-o_t},\widehat{\Delta x},\widehat{\Delta y},
\widehat{\Delta\psi},\widehat P,\widehat h_{\min},
\widehat{\theta}_{\max},\widehat{\omega}_{\max}),
\]
where one macro step is four low-level steps ($0.08$\,s). The model has two
hidden layers of width 512 and is trained on 42{,}054 windows from 207 episode
groups with disjoint train/validation/test groups. Each ensemble member
recursively advances its own predicted full observation residual; the command
slice is overwritten by the next planned command before the following query.
At four macro steps ($0.32$\,s), held-out open-loop errors are $3.3$\,cm
displacement MAE, $5.17$\,J cumulative-work MAE, $8.8$\,mm minimum-height MAE,
and $0.756$ normalized full-state RMSE.

MPC enumerates all $3^4=81$ scale sequences from
$\{0.75,0.90,1.00\}$ and executes the first command of the selected sequence.
After the four-step rollout, a second, separately trained task-value network
predicts remaining work, remaining time, and success under direct-control
continuation from the predicted terminal observation and local target. It is
trained on 4{,}298 samples from 144 mixed-controller episodes; held-out metrics
are work MAE $15.7$\,J ($R^2=0.846$), time MAE $0.362$\,s
($R^2=0.654$), and success AUC $0.819$. MPC minimizes ensemble-upper-bound total
work subject to predicted success, time, posture, and optional stability
constraints, with exact direct-sequence fallback.

Thus MPC shares the decision rate and command bounds with VGCC but not its model
or data budget. Its 81 sequence evaluations are numerically close to VGCC's
approximately 70 one-step candidates, but each sequence uses four recursive
ensemble transitions plus a terminal-value query. We describe it as a heavier
model-based baseline, not a same-model or equal-compute comparison.

\textbf{Mixed-terrain evaluation (Section~\ref{sec:results-headroom}).}
\label{app:hetero}
The flat and rough single-type tasks replace the terrain generator of the
otherwise unchanged rough environment with a single sub-terrain each: a plane,
and uniform height noise of $2$--$10$\,cm; every scene, observation, command,
and event setting is identical, and the same pretrained checkpoint is used.
The response model for these deployments is identified once on pooled excitation
from both terrains with the same protocol ($\approx$25k windows; displacement
$R^2=0.92$, cost $0.73$). Fixed scales $\{0.60, 0.75, 0.90, 1.00\}$ and VGCC run
from identical initial states on three seeds disjoint from all tuning. Because
the world-frame final-height criterion is dominated by stop-point ground elevation
on $\pm10$\,cm noise, completion means reaching the target without early
termination. Table~\ref{tab:hetero} reports the paired frontier: COT falls
monotonically on both terrains. The budget analysis pairs one flat with one rough
leg, sweeps a shared completion-time budget over the feasible range, and compares
the best budget-feasible uniform scale against the best per-terrain scale
allocation on the convex hulls of the per-terrain (work, time) means; the
reported interval is an episode bootstrap over both legs.

\begin{table}[h]
\centering
\caption{Single-terrain-type paired evaluation (Go2, three seeds, eight targets $\times$ two trials; completion = target reached without early termination; flat $n=48$, rough $n=36$ jointly completed episodes per scale, of 48 evaluated). COT falls monotonically with the scale on both terrains.}
\label{tab:hetero}
\begin{tabular}{lcccccc}
\toprule
& \multicolumn{3}{c}{Flat} & \multicolumn{3}{c}{Rough ($2$--$10$\,cm)} \\
\cmidrule(lr){2-4}\cmidrule(lr){5-7}
Scale & COT & Work (J) & Time (s) & COT & Work (J) & Time (s) \\
\midrule
1.00 & 1.027 & 179.6 & 1.80 & 1.670 & 290.5 & 2.16 \\
0.90 & 0.964 & 166.7 & 1.99 & 1.598 & 278.5 & 2.30 \\
0.75 & 0.897 & 152.6 & 2.36 & 1.513 & 265.2 & 2.76 \\
0.60 & 0.882 & 148.8 & 2.90 & 1.447 & 247.8 & 3.31 \\
\bottomrule
\end{tabular}
\end{table}

\textbf{Archived exact-replay scale-prefix analysis.}
The Go2 analysis of Section~\ref{sec:results-headroom} uses ten source seeds, 48 query
states per seed (480 total), and constant four-macro-step prefixes from
$\{0.75,0.90,1.00\}$. For each query and scale, the simulator is reset to the
episode start, the recorded direct-control prefix is replayed to the query, the
candidate prefix is executed, and direct control completes the episode. The
terrain curriculum is frozen; all query observations reconstruct exactly
(maximum observation error zero). This start-state replay is necessary because
Isaac Lab's scene-state reset does not restore policy observation and action
history. Five source seeds were collected initially, followed by five additional
seeds under the same protocol. The descriptive four-way feature ablation below
uses the initial subset; the main text reports the observation-only and
observation+$F_\eta$ comparisons on all ten seeds.

\textbf{Action-aligned confirmatory protocol.}
For Go2, seeds 1301--1320 define twenty independent source clusters for each of
three separately collected query-state distributions: direct control, evaluated
VGCC, and evaluated MPC. Each cluster contributes 48 queries, yielding 960
queries per distribution. The branch set retains the direct command and scales
$0.75$ and $0.90$, and adds heading offsets $\pm0.3$ rad and yaw-rate offsets
$\pm0.2$ rad/s. Candidate interventions pass through the deployed bounded blend
($\alpha=0.5$) and execute for four macro steps. The complete realized command
history is replayed before every query, and all seven branches begin from the
same reconstructed observation. Matched-frequency and cross-fitted fixed
comparators are computed separately within each query distribution.

The confirmatory learner uses observation-only features, thereby avoiding
dependence on an auxiliary predictive model at audit time. Each of 200
source-cluster bootstrap replicates refits the treatment-effect learner and
evaluates all estimands on out-of-bootstrap clusters. The cross-fitted fixed
reference is selected using training clusters only. Table~\ref{tab:confirmatory-audit}
reports the resulting three-channel decisions.

\textbf{H1 query-distribution correction.}
The earlier H1 pilot sampled query states from episodes controlled by
dimension-matched, randomly initialized macro-transition and task-value
surrogates. Although every branch reconstructed its query observation exactly,
the surrogates altered the distribution of visited states. Because that
distribution is part of the audit estimand, the four-cluster pilot remains
excluded. The confirmatory collection instead uses seeds 1401--1420, samples all
pre-intervention states under direct control (scale $1.0$), loads neither
surrogate, and evaluates the same seven interventions at 48 queries per cluster
(960 total). Its twenty independent clusters and 200 complete learner refits
support the formal H1/direct decision in
Table~\ref{tab:confirmatory-audit}.

The same-state oracle requires candidate success no worse than the direct branch
and total remaining time no greater than $1.10$ times direct, then minimizes
total mechanical work. The treatment-effect model is a five-member, two-layer
MLP ensemble (hidden width 64) trained for 200 epochs. Inputs contain the
deployment observation, local target, candidate-intervention encoding, and, for
the archived scale analysis, differences between
candidate and direct macro-response ensemble predictions (work, progress,
minimum height, maximum tilt, and maximum angular speed; means and
uncertainties). Targets are candidate work and time relative to the direct
branch and realized eligibility. Evaluation leaves one entire source seed out.
The selector accepts a candidate only when its work upper confidence estimate is
negative, its time upper estimate meets the budget, and its eligibility
probability lower estimate is at least $0.5$.

For each held-out fold, the matched randomized mixture uses the selector's exact
marginal action frequencies but assigns them
independently of query state. Its reported metrics are the exact expectation over
the replayed potential outcomes, so no Monte Carlo sampling noise is introduced.
The selector--mixture difference isolates state-dependent allocation from the
gain attributable solely to changing action composition. This is a post-hoc
attribution comparator, not a deployable policy. Separately, the cross-fitted
fixed reference is the lowest-work fixed action satisfying the violation
tolerance on training clusters only and is evaluated on the held-out cluster.
Confirmatory intervals resample whole source clusters, refit the learner, and
evaluate on out-of-bootstrap clusters.

\begin{table}[h]
\centering
\caption{Diagnostics for the action-aligned confirmatory audit. Activation is
the fraction of queries on which the learned selector departs from direct
control; agreement is with the same-state oracle. The cross-fitted fixed action
is selected using training clusters only.}
\label{tab:confirmatory-diagnostics}
\begin{tabular}{lcccc}
\toprule
Embodiment/query distribution & Work MAE & Activation & Oracle agreement & Cross-fitted fixed action \\
\midrule
Go2/direct & $5.90\%$ & $33.13\%$ & $16.46\%$ & yaw $-0.2$ \\
Go2/VGCC & $5.92\%$ & $38.85\%$ & $16.15\%$ & direct \\
Go2/MPC & $6.54\%$ & $33.75\%$ & $13.23\%$ & yaw $-0.2$ \\
H1/direct & $4.62\%$ & $30.00\%$ & $13.85\%$ & scale $0.90$ \\
\bottomrule
\end{tabular}
\end{table}

\subsection{Decision-Boundary Resolution Diagnostic}
\label{app:audit-resolution}

To distinguish sampling uncertainty from an unfavorable point estimate, we
compute a post-hoc resolution diagnostic for the three confirmatory channels.
Let $N=20$, let $w$ be the directional distance from the point estimate to its
current limiting 95\% bound, and let $m$ be the distance from the point estimate
to the corresponding decision threshold. Under the planning approximation
$w_N\propto N^{-1/2}$, the cluster count at which that bound reaches the
threshold is
\[
\widetilde N=\left\lceil N\left(\frac{w}{m}\right)^2\right\rceil.
\]
The calculation is defined only when the point estimate lies on the
\textsc{Go} side of the threshold. If it does not, variance reduction around an
unchanged estimate cannot satisfy that channel. This approximation ignores
changes in the fitted learner and non-Gaussian tail behavior; it is therefore a
design aid, not a prospective power calculation or a revision of the nested-refit
decision.

\begin{table}[h]
\centering
\caption{Approximate clusters required for each limiting confidence bound to
reach the headline \textsc{Go} boundary under fixed point estimates and
$N^{-1/2}$ directional-width scaling. ``Point below'' means the gain estimate is
itself below $1\%$, so interval narrowing alone cannot satisfy that channel.
Formal decisions remain those in Table~\ref{tab:confirmatory-audit}.}
\label{tab:audit-resolution}
\begin{tabular}{lccc}
\toprule
Embodiment/query distribution & Deployment lower bound & Allocation lower bound & Violation upper bound \\
\midrule
Go2/direct & point below & point below & 28 \\
Go2/VGCC & 86 & point below & 97 \\
Go2/MPC & point below & point below & 42 \\
H1/direct & point below & point below & 20 (already satisfied) \\
\bottomrule
\end{tabular}
\end{table}

\subsection{Treatment-Effect Feature Ablation}
\label{app:feature-ablation}

We isolate which deployment-time representation contributes to the final
decision while holding the treatment-effect ensemble, five leave-one-source-seed-out
folds, confidence rule, and matched-mixture evaluation fixed. All variants
receive the current observation summary, local target, and candidate scale.
The $f_\phi$ variant adds candidate-versus-direct means and ensemble
uncertainties from the six-output one-window response model. The $F_\eta$
variant adds the corresponding four-step macro-prefix predictions used by MPC.

\begin{table}[h]
\centering
\caption{Descriptive feature ablation for the exact-replay treatment-effect selector (240 queries). ``Allocation gain'' is selector work relative to its own frequency-matched randomized mixture; negative is better. The prespecified main representation is observation+$F_\eta$.}
\label{tab:feature-ablation}
\begin{tabular}{lccccc}
\toprule
Features & Work MAE & Active & Work vs.\ direct & Allocation gain & Violations \\
\midrule
observation only & $12.60\%$ & $8.8\%$ & $-0.43\%$ & $-0.38\%$ & $0.42\%$ \\
observation + $f_\phi$ & $12.93\%$ & $13.8\%$ & $-0.37\%$ & $-0.35\%$ & $4.17\%$ \\
observation + $F_\eta$ & $\mathbf{12.22\%}$ & $12.9\%$ & $-0.54\%$ & $-0.42\%$ & $1.67\%$ \\
observation + $f_\phi+F_\eta$ & $12.33\%$ & $12.1\%$ & $\mathbf{-0.76\%}$ & $\mathbf{-0.76\%}$ & $2.92\%$ \\
\bottomrule
\end{tabular}
\end{table}

Observation-only features recover nearly the same allocation gain as the
prespecified observation+$F_\eta$ representation ($0.38\%$ versus $0.42\%$).
Adding $f_\phi$ alone does not improve work error or allocation and increases
realized constraint violations. Combining both identified models gives the
largest point estimate, $0.76\%$, but this interaction was discovered on the
same five folds, is below one percent, and incurs more violations; it is not a
nested or independently confirmed model-selection result. We therefore retain
observation+$F_\eta$ as the main prespecified audit and report the combined row
only as a hypothesis for future confirmation. The robust conclusion is that
the treatment-effect ensemble extracts most of its limited signal from the
observation itself; neither identified model provides a decisive incremental
contribution on this dataset.

\subsection{Audit Calibration Controls}
\label{app:audit-controls}

The semi-synthetic calibration changes only candidate mechanical-work outcomes;
observations, targets, success, time, posture, source-seed splits, and the
treatment-effect learner remain unchanged. For observable controls, the
preferred scale is $0.90$ when the pre-intervention local target has
$y\ge0$ and $0.75$ otherwise, and its realized work is multiplied by
$1-\gamma$. Hidden controls draw the preferred scale independently for each
query using a deterministic seed and do not expose that draw in any feature.
The matched-mixture calculation is repeated after each modification.

Confidence intervals resample the ten source-seed folds as clusters with
100{,}000 bootstrap replicates (seed 20260717). Table~\ref{tab:control-curve}
reports the ten-seed observable-signal sweep used in the main text. These
intervals describe the archived allocation and violation channels; they do not
include the cross-fitted-fixed deployment channel or learner refitting.

\begin{table}[h]
\centering
\caption{Descriptive observable-signal calibration on ten source seeds.}
\label{tab:control-curve}
\begin{tabular}{lcc}
\toprule
$\gamma$ & $H_{\rm alloc}$, 95\% CI & Mean $q$ \\
\midrule
$2\%$  & $0.53\ [0.17,0.87]\%$ & $3.12\%$ \\
$5\%$  & $1.26\ [0.75,1.78]\%$ & $4.58\%$ \\
$9\%$  & $2.32\ [1.67,2.98]\%$ & $5.42\%$ \\
$10\%$ & $2.61\ [1.85,3.30]\%$ & $5.42\%$ \\
\bottomrule
\end{tabular}
\end{table}

Relative to the five-seed analysis, estimated allocation gain increases for
$\gamma\ge5\%$, but the violation intervals also widen across $\kappa=5\%$.
This sensitivity illustrates the importance of cluster-level uncertainty and
adequate numbers of independent units; no formal decision is assigned to this
archived allocation-and-violation sweep.

To estimate specificity, we repeat the $5\%$ hidden-signal control with ten
independent mappings on the same ten source seeds. The allocation lower bound
exceeds $0.25\%$, $0.50\%$, and $1.00\%$ in $3/10$, $2/10$, and $0/10$
mappings, respectively. This motivates retaining $1\%$ as the present
allocation threshold, but does not establish a universal constant: another
domain requires its own deployment value and control-based calibration.

The learner-level positive control comprises twenty synthetic source clusters.
Target sign identifies which of two non-direct actions receives a $20\%$ work
benefit, while both candidates satisfy the success and time constraints. The
control executes treatment-effect training, conservative selection, the
matched-frequency and cross-fitted fixed comparisons, cluster-bootstrap
refitting with out-of-bootstrap evaluation, and Eq.~\ref{eq:audit-decision}.
With $\delta_{\rm dep}=\delta_{\rm alloc}=1\%$ and $\kappa=5\%$, it returns
\textsc{Go}. This verifies the complete analytical procedure rather than only
the Boolean decision logic.

\section{Secondary Experiments and Diagnostics}
\label{app:secondary}

This section reports complementary diagnostic evidence: qualitative and
per-target results, a retrieval-based alternative, and the complete component
ablation.

\subsection{Qualitative and Per-Target Results}
\label{app:qualitative}

\begin{figure}[h]
\centering
\includegraphics[width=\linewidth]{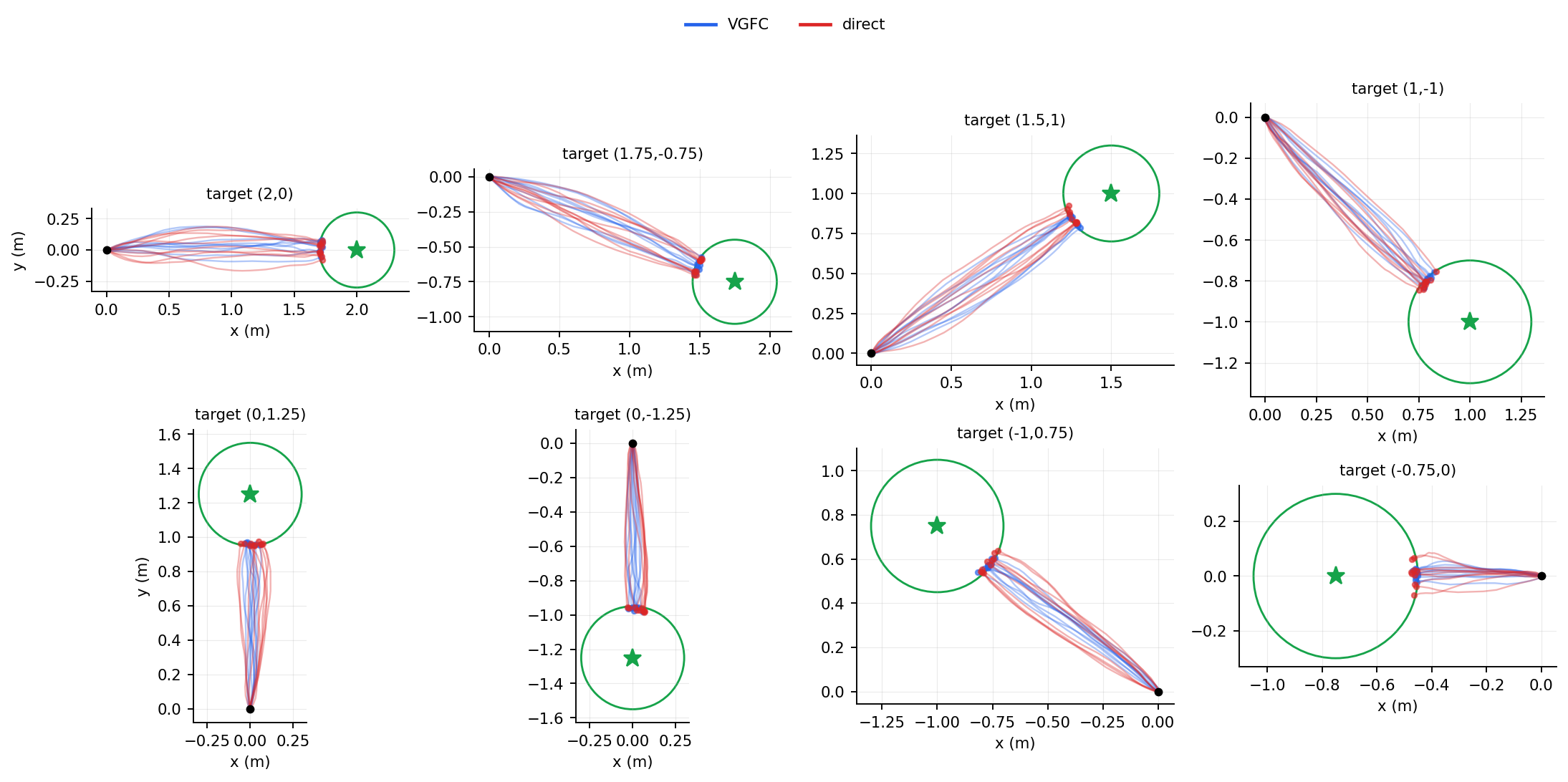}
\caption{XY paths on the eight harder targets (one evaluation seed, 10 trials per method). VGCC and direct control reach similar target neighborhoods.}
\label{fig:traj}
\end{figure}

\begin{figure}[h]
\centering
\includegraphics[width=\linewidth]{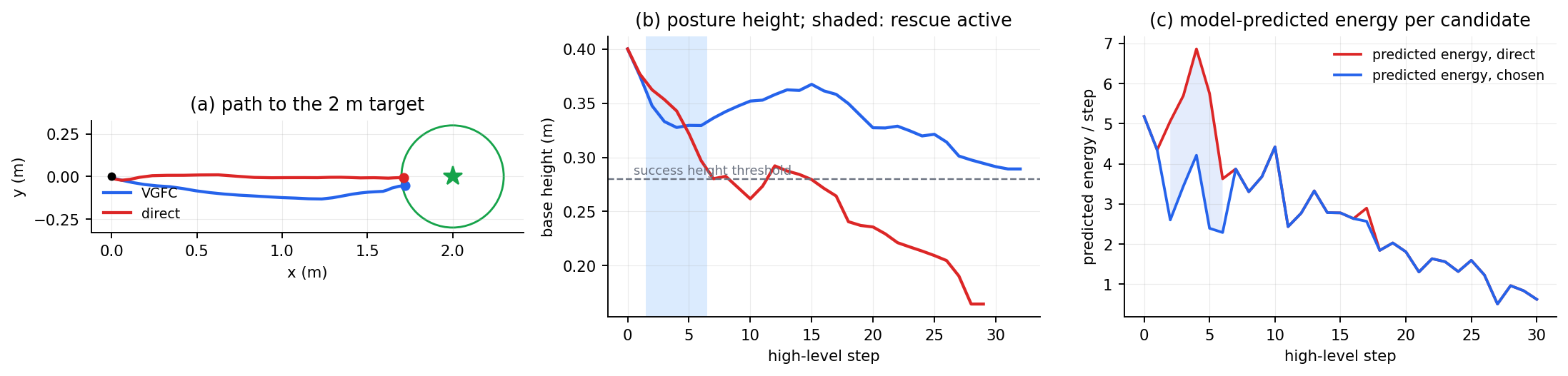}
\caption{Selected paired episode illustrating rescue behavior. This example shows mechanism, not prevalence.}
\label{fig:example}
\end{figure}

The selected example forecasts a posture-floor breach and activates rescue. At
the aggregate level, the principal residual error is arrival with low posture. A symmetric
posture-aware stopping rule did not improve success at $n=400$ per method; it
instead used the remaining horizon for recovery attempts. The example is
therefore interpreted as a mechanism illustration rather than evidence of a
success-rate effect; aggregate evidence is
in Tables~\ref{tab:physical-energy} and~\ref{tab:coverage}.

Table~\ref{tab:pertarget} gives the per-target breakdown of the harder benchmark.
The proxy reduction holds on seven of eight targets, but success differences change
sign at $n=50$ per target, which is why we report aggregate success rather than a
target-specific success effect.

\begin{table}[h]
\centering
\caption{Per-target harder-target results (50 episodes per target and method), ordered by proxy reduction.}
\label{tab:pertarget}
\begin{tabular}{lccccccc}
\toprule
& \multicolumn{3}{c}{Effort proxy $\downarrow$} &
\multicolumn{2}{c}{Success $\uparrow$} & \multicolumn{2}{c}{Final height (m)} \\
\cmidrule(lr){2-4}\cmidrule(lr){5-6}\cmidrule(lr){7-8}
Target & VGCC & direct & $\Delta$ & VGCC & direct & VGCC & direct \\
\midrule
$(0,1.25)$ & 1.815 & 2.083 & $-12.9\%$ & 0.940 & 0.980 & 0.340 & 0.351 \\
$(1.75,-0.75)$ & 2.217 & 2.542 & $-12.8\%$ & 0.480 & 0.520 & 0.276 & 0.285 \\
$(-1,0.75)$ & 1.861 & 2.122 & $-12.3\%$ & 1.000 & 1.000 & 0.358 & 0.367 \\
$(0,-1.25)$ & 1.822 & 2.073 & $-12.1\%$ & 0.940 & 0.980 & 0.345 & 0.350 \\
$(1.5,1)$ & 2.203 & 2.471 & $-10.8\%$ & 0.800 & 0.740 & 0.316 & 0.313 \\
$(2,0)$ & 2.367 & 2.619 & $-9.6\%$ & 0.240 & 0.260 & 0.233 & 0.224 \\
$(1,-1)$ & 2.028 & 2.240 & $-9.5\%$ & 0.980 & 0.960 & 0.355 & 0.349 \\
$(-0.75,0)$ & 1.496 & 1.510 & $-0.9\%$ & 1.000 & 1.000 & 0.359 & 0.363 \\
\bottomrule
\end{tabular}
\end{table}

\subsection{Retrieval Alternative (VGSR)}
\label{app:retrieval}

A nonparametric alternative to the response model retrieves reusable
behavior chunks instead of predicting command responses. VGSR segments
converged-policy rollouts into chunks, describes them by body-frame displacement,
velocity, yaw, posture, effort proxy, stability, and smoothness, then
novelty-filters and clusters them into a behavior archive. At deployment it
retrieves a chunk by progress, state applicability, effort, stability, and utility,
and blends the chunk velocity with the direct command only when a heuristic gate
accepts it. Unlike VGCC, it can interpolate only inside archive support.

On the original five-target protocol, VGSR matches VGCC's proxy level on the harder
family (2.061 versus 2.073) but has 1.2\% terminations and lower final height. It
\emph{increases} the proxy by 3.8\% on held-out directions and 4.4\% on H1, exactly
the generalization the parametric model provides and the archive does not. In a
separate five-seed harder-target aggregate it reduces the proxy by 7.8\% with
success 0.785 versus 0.800. Removing its gate increases the proxy by more than
50\%. Archive size is non-monotonic: the top 2,000 chunks improve success relative
to all 3,904, while top-500 minimizes proxy at lower success. These coverage and
curation dependencies motivate the parametric response model used by VGCC.

\subsection{Complete VGCC Component Ablation}
\label{app:ablation}

\begin{table}[h]
\centering
\caption{Paired component ablations on two seeds ($n=80$ per variant). Differences are against direct control in the same runs.}
\label{tab:vgcc-ablation}
\begin{tabular}{lccc}
\toprule
Variant & $\Delta$ proxy $\downarrow$ & $\Delta$ success & Active steps \\
\midrule
VGCC (full) & $-12.2\%$ & $+0.012$ & 11.7 \\
no annealing & $-12.2\%$ & $+0.013$ & 12.3 \\
no viability gate & $-4.8\%$ & $-0.025$ & 8.9 \\
no progress floor & $-19.6\%$ & $-0.037$ & 19.8 \\
substitution ($\alpha=1$) & $-14.0\%$ & $-0.050$ & 16.3 \\
25\% identification data & $-6.4\%$ & $-0.012$ & 10.7 \\
\bottomrule
\end{tabular}
\end{table}

The gate restricts predicted savings to the model's empirically supported posture regime;
the progress floor limits behavioral drift; and bounded blending avoids the larger
success cost of substitution. A quarter of the identification data retains
displacement $R^2=0.92$ but reduces the cost-channel $R^2$ from 0.75 to 0.62,
halving the realized proxy reduction. The absolute proxy margin does not transfer
across embodiments: it increases the H1 proxy by 13.5\%. This result motivates
the relative margin in Eq.~\ref{eq:vgc-cost}.

\textbf{Objective ablation.}
A variant that ranks candidates by predicted cost \emph{per predicted progress}
(with execution-aligned scoring and all other mechanisms unchanged) was evaluated
against the standard gate under the prespecified frontier protocol (three seeds,
$n=48$ episodes per method, identical initial states). It forfeits most of the
compensation: effort proxy $2.010$ versus $1.810$ for the raw-cost gate (direct
$2.132$), mechanical work $132.3$ versus $126.9$\,J, and success $0.792$ versus
$0.854$, and it intervenes less (elapsed time $2.08$ versus $2.25$\,s). The
per-seed paired proxy difference favors the raw-cost gate on all three seeds.
Because the progress floor already limits slowdown, normalization by predicted
progress imposes an additional penalty on slower candidates and rejects some
interventions that otherwise satisfy the constraint. These results support
representing slowdown through the progress constraint rather than through both
the constraint and the objective.

\end{document}